\documentclass[]{fairmeta}
\makeatletter
\newcommand{\authorbreak}{%
  \g@addto@macro\authorlist{\\[-0.05em]}%
  \def\authorlistsep{}%
}
\makeatother

\usepackage{hyperref}
\usepackage{url}
\usepackage{algorithm}
\usepackage{algpseudocode}
\usepackage{multirow}
 \usepackage[normalem]{ulem}   %
\algrenewcommand\algorithmicrequire{\textbf{Input:}}
\algrenewcommand\algorithmicensure{\textbf{Output:}}
\usepackage{dblfloatfix}

\usepackage{hyperref}
\usepackage{url}
\usepackage{booktabs}
\usepackage{titletoc}
\usepackage{caption}
\usepackage{capt-of}
\crefname{algorithm}{Alg.}{Algs.}
\Crefname{algorithm}{Alg.}{Algs.} 
\crefname{section}{Sec.}{Secs.}
\Crefname{section}{Sec.}{Secs.}
\usepackage[table]{xcolor}
\definecolor{fklblue}{HTML}{4C78A8}
\definecolor{rklred}{HTML}{9B1B30}
\definecolor{sdgold}{HTML}{E3B23C}
\definecolor{studentpurple}{HTML}{7A3FB0}

\usepackage{colortbl}
\usepackage{amssymb}
\usepackage{graphicx}
\newcommand{\blackdiamond}{\rotatebox[origin=c]{45}{$\scriptstyle\blacksquare$}}
\usepackage{float}
\usepackage{comment}
\newtoggle{comment}
\toggletrue{comment}
\usepackage{etoolbox} 

\newcommand{\ours}{\textrm{\textsc{Switch Distillation}}}
\newcommand{\olmotwo}{OLMo-2}
\newcommand{\tinycite}[1]{{\scriptsize\citep{#1}}}

\crefformat{section}{§#2#1#3}
\crefformat{subsection}{§#2#1#3}
\crefformat{subsubsection}{§#2#1#3}
\newcommand{\appref}[1]{\hyperref[#1]{App.~\ref*{#1}}}
\title{Knowledge Distillation During Mid-Training \\
Favors Reasoning over Factual Recall}

\renewcommand{\authorlist}{%
  \begin{tabular}{@{}l@{}}
    \authorformat[1,2]{Jacqueline He},
    \authorformat[3]{Howard Yen},
    \authorformat[1,2]{Shuyue Stella Li},
    \authorformat[2]{Margaret Li},
    \\[2pt]
    \authorformat[1]{Hanqing Zeng},
    \authorformat[1]{Yinglong Xia},
    \authorformat[1]{Benyu Zhang},
    \authorformat[1]{Zhuokai Zhao},  \\[2pt]
    \authorformat[1]{Qiang Zhang},
    \authorformat[2]{Pang Wei Koh},
    \authorformat[1,2]{Luke Zettlemoyer},
    \authorformat[1]{Wen-tau Yih}
  \end{tabular}%
}

\affiliation[1]{Meta AI}
\affiliation[2]{University of Washington}
\affiliation[3]{Princeton University}

\abstract{
\vspace{-1mm}
Logit-based knowledge distillation (KD) is used to train smaller language models (LMs) via supervision from stronger teachers, but whether its benefits are consistent across training stages remains unclear.
Through controlled experiments, we find that forward Kullback-Leibler (KL) distillation---the standard KD formulation---with post-trained teachers behaves fundamentally differently during mid-training, an intermediate phase of self-supervised learning on curated corpora.
Surprisingly, while forward KD simultaneously improves reasoning and factual recall during pre-training relative to standard next-token prediction (NTP), it instead slows factual recall acquisition during mid-training despite continued reasoning gains.
We trace this stage dependence to an asymmetry in teacher confidence across data domains and the student’s evolving knowledge state: teachers are more confident on procedural than unstructured text data, while students acquire low-entropy factual knowledge earlier in training.
To mitigate this imbalance, we propose \ours{}, a simple mid-training objective that distills on tokens where the teacher is confident, using teacher \emph{predictive entropy} as a lightweight routing signal, and otherwise falls back to cross-entropy.
\ours{} consistently outperforms existing distillation objectives across teacher sizes. Relative to NTP, it achieves 1.61–1.71$\times$ the reasoning performance and 1.13–1.19$\times$ the knowledge and commonsense performance while preserving 96.7--96.8\% of factual recall.
Crucially, these benefits persist after post-training: \ours{} closes the factual recall gap while maintaining 1.25–1.32$\times$ and 1.13–1.20$\times$ gains in reasoning and knowledge and commonsense, respectively.

}

\date{\today}
\correspondence{Jacqueline He at \email{jyyh@cs.washington.edu}}

\metadata[Code]{\url{https://github.com/facebookresearch/midtraining-distillation}}

\begin{document}

\maketitle

\begin{figure*}[htbp]
    \centering
    \includegraphics[width=\textwidth]{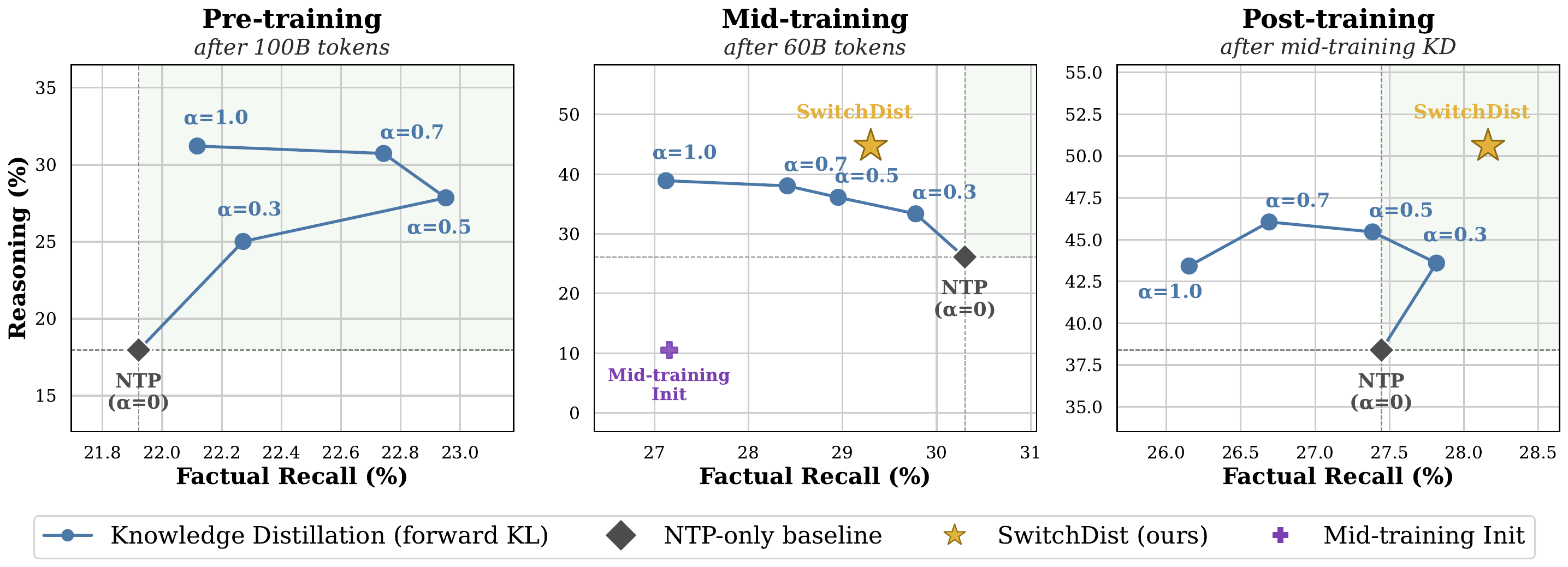}
    \vspace{-16pt}
    \caption{
 \textbf{Knowledge distillation (KD) generally improves reasoning, but its effect on factual recall changes across training stages.} 
 Using an OLMo-2 1B student and 7B-Instruct teacher, we sweep the distillation strength $\alpha$, which interpolates between NTP ($\alpha=0$) and \textcolor{fklblue}{forward-KL} distillation.
 Distillation improves both reasoning and factual recall during pre-training (left), but favors reasoning at the expense of factual recall during mid-training (middle). \textcolor{sdgold}{\ours{}} (\textcolor{sdgold}{\raisebox{-0.1ex}{$\bigstar$}}) mitigates this imbalance and Pareto-dominates NTP after post-training (right).}
    \label{fig:teaser}
\end{figure*}
\section{Introduction}
Modern language models (LMs) increasingly rely on a dedicated \emph{mid-training} stage between pre-training and post-training, in which self-supervised next-token prediction continues on a smaller, high-quality corpus curated to improve capabilities such as factuality, reasoning, coding, and instruction following~\citep{grattafiori2024llama3herdmodels, allal2025smollm, walsh2025,liu2026midtrainingbridgespretrainingposttraining, meta2026glimmer}. 
Because mid-training uses far fewer tokens than pre-training, extracting more learning signal from each token becomes especially important. Knowledge distillation (KD)~\citep{10.1145/1150402.1150464, hinton2015distillingknowledgeneuralnetwork} offers a natural approach by augmenting ground-truth next-token supervision with the richer predictive distribution of a stronger teacher, typically through minimizing the forward Kullback-Leibler divergence between teacher and student predictive distributions~\citep{gu2024minillm, zhong-etal-2024-revisiting}. Yet despite its growing use in frontier language modeling pipelines~\citep{gemmateam2025gemma3technicalreport, meta2026glimmer}, KD has been studied almost exclusively in pre-training and post-training~\citep{busbridge2025distillation, lu2026strongteacherneededdistillation, agarwal2024onpolicydistillationlanguagemodels}, leaving it unclear whether its benefits transfer to the relatively nascent mid-training stage.

Surprisingly, we find that knowledge distillation behaves qualitatively differently in mid-training than in pre-training. Using the \olmotwo{} ecosystem~\citep{walsh2025}, one of the most recent fully open model families with intermediate checkpoints, training recipes, and multiple model scales, we conduct controlled pre-training and mid-training experiments with 1B students. As \cref{fig:teaser} shows, increasing distillation strength ($\alpha$) generally improves reasoning across training stages, but its effect on factual recall differs sharply.
While traditional forward-KL distillation improves both reasoning and factual recall over vanilla next-token prediction (NTP) during pre-training, its reasoning gains are accompanied by comparatively lower factual recall during mid-training. 
We refer to this phenomenon as the \emph{reasoning–recall tradeoff}.

This tradeoff is remarkably robust: across instruction-tuned teacher sizes, KL directions, and interpolation coefficients, no KD objective Pareto-dominates NTP during mid-training. We track this behavior to an interaction between teacher confidence, the student's evolving knowledge state, and the distillation objective. 
To begin, our teachers exhibit substantially lower predictive entropy on procedural data, such as math and instruction-following, than on unstructured data such as general web text; lower entropy is also strongly correlated with higher-quality supervision.
Students acquire factual knowledge associated with lower teacher entropy earlier during pre-training, such that facts that have not been learned at the start of mid-training receive disproportionately weak teacher supervision.
Finally, as teacher entropy rises, KD increasingly attenuates the ground-truth learning signal relative to NTP.
Together, these effects provide evidence for why distillation preferentially accelerates reasoning while reducing open-ended factual recall relative to NTP during mid-training. 

Motivated by this analysis, we propose \ours{}, a drop-in mid-training objective that uses teacher predictive entropy to route each token between distillation and next-token prediction. Because this routing signal is computed from the teacher logits already required for KD, \ours{} incurs minimal additional computation. 
Across 7B and 13B teachers, \ours{} substantially improves the reasoning–recall tradeoff.
With an \olmotwo{} 7B Instruct teacher, for example, \ours{} improves average  reasoning performance by 71\% and knowledge and commonsense performance by 19\% over NTP, while reducing factual recall by just 1 percentage point. Given that the purpose of mid-training is to provide a strong prior for alignment, we show that \ours{}'s gains persist through post-training: reasoning remains 32\% higher, knowledge tasks improve by 20\%, and the factual recall gap closes entirely.
Our contributions are threefold:
\begin{enumerate}
\item \textbf{Empirical finding:} We uncover a robust reasoning--recall tradeoff during the mid-training regime: distillation of a substantially pre-trained student improves reasoning while slowing factual recall relative to NTP. 

\item \textbf{Explanatory analysis:} We explain this tradeoff through the interaction between teacher confidence, student learning dynamics, and the distillation objective, showing that facts not yet learned by the pre-trained student disproportionately receive weak teacher supervision.

\item \textbf{Mid-training objective:} Our analysis naturally motivates \ours{}, which routes tokens between KD and NTP using teacher predictive entropy. Our method substantially mitigates the tradeoff across teacher sizes and retains its gains after post-training.
\end{enumerate}

\section{Background}
\subsection{Preliminaries}
\paragraph{Language modeling.}
Given a token sequence $\mathbf{x}=(x_1,\ldots,x_N)$ of length $N$ and an auto-regressive language model distribution $p_\theta$, standard next-token prediction minimizes the expected cross-entropy loss
\begin{align} \mathcal{L}_{\mathrm{CE}} = \mathbb{E}_{(\mathbf{x},n)} \left[ -\log p_\theta(x_n \mid x_{<n}) \right],
\end{align}
where $x_n$ is the target next token and the expectation is over training sequences and token positions.

\paragraph{Knowledge distillation.}
Logit-based knowledge distillation (KD) trains a student to match the output distribution of a vocabulary-compatible teacher~\citep{hinton2015distillingknowledgeneuralnetwork,busbridge2025distillation}: 
\begin{align}
\mathcal{L}_{\mathrm{KD}}
=
(1-\alpha)\mathcal{L}_{\mathrm{CE}}
+
\alpha\mathcal{L}_{\mathrm{KL}},
\end{align}
where $\alpha\in[0,1]$ controls the distillation strength. 
Let $p_T^{(\tau)}(\cdot\mid x_{<n})$ and $p_S^{(\tau)}(\cdot\mid x_{<n})$ denote the next-token distributions scaled by temperature $\tau>0$ for a teacher $T$ and student $S$, respectively.  
Standard KD instantiates $\mathcal{L}_{\mathrm{KL}}$ using the forward KL (FKL) divergence~\citep{Kullback51klDivergence}:
\begin{align}
\mathcal{L}_{\mathrm{FKL}}
&= \tau^2\,\mathbb{E}_{(\mathbf{x},n)}
\left[\sum_{v\in\mathcal{V}} p_T^{(\tau)}(v\mid x_{<n})
\log\frac{p_T^{(\tau)}(v\mid x_{<n})}{p_S^{(\tau)}(v\mid x_{<n})}\right].
\end{align}
Recent work has advocated for distillation using the reverse KL (RKL) divergence, which discourages student mass on the teacher's low-probability regions~\citep{agarwal2024onpolicydistillationlanguagemodels,gu2024minillm}:
\begin{align}
\mathcal{L}_{\mathrm{RKL}}
&= \tau^2\,\mathbb{E}_{(\mathbf{x},n)}
\left[\sum_{v\in\mathcal{V}} p_S^{(\tau)}(v\mid x_{<n})
\log\frac{p_S^{(\tau)}(v\mid x_{<n})}{p_T^{(\tau)}(v\mid x_{<n})}\right].
\end{align}
Throughout this paper, we instantiate $\mathcal{L}_{\mathrm{KL}}$ as either $\mathcal{L}_{\mathrm{FKL}}$ or $\mathcal{L}_{\mathrm{RKL}}$, corresponding to forward-KL distillation (FKD) and reverse-KL distillation (RKD), respectively.

\subsection{Experimental Setup}
\label{subsec:experimental}
\paragraph{Training regimes.} We build on the open-source \olmotwo{} ecosystem~\citep{walsh2025}. We pre-train and mid-train on Dolmino Mix 1124, a data mixture of filtered DCLM web text, FLAN instruction-following data, Dolmino Math, peS2o, Wikipedia (including Wikibooks), and Stack Exchange. For pre-training, we initialize 1B students from random weights and train beyond Chinchilla optimality for 100B tokens~\citep{hoffmann2022trainingcomputeoptimallargelanguage}. For mid-training, we initialize from the \olmotwo{} 1B Stage 1 checkpoint, already pre-trained on 4T tokens, and continue training for 60B tokens.

\paragraph{Teacher models.} Post-trained models are increasingly used as teachers for reference-based language modeling in recent research~\citep{goyal2026distilled, huang2026remitrlguidedmidtrainingiterative, jin2026entropyawareonpolicydistillationlanguage, tan2026selfimprovingpretrainingusingposttrained} and frontier LLM pipelines~\citep{gemmateam2025gemma3technicalreport, meta2026glimmer}. Their stronger instruction-following and reasoning abilities make them natural choices for capability transfer. Accordingly, we employ \olmotwo{} 1B Instruct, 7B Instruct, and 13B Instruct as teachers.

\paragraph{Evaluation.} We evaluate with the standardized OLMES~\citep{gu-etal-2025-olmes} harness, grouping benchmark tasks into \textsc{Reasoning} (generative problem solving), \textsc{Factual Recall} (open-ended generative retrieval of factual knowledge), \textsc{Knowledge \& Commonsense} (multiple choice world knowledge and commonsense reasoning), and \textsc{Instruction Following} (post-training only). We report macro-averages of each task group and focus our stage-dependent analysis on \textsc{Reasoning} and \textsc{Factual Recall}. See \cref{tab:evaluation_tasks} for full task suite and evaluation settings.

\section{Characterizing Mid-Training Distillation}
\label{sec:observation}
\begin{figure*}[t]
    \centering
    \includegraphics[width=\textwidth]{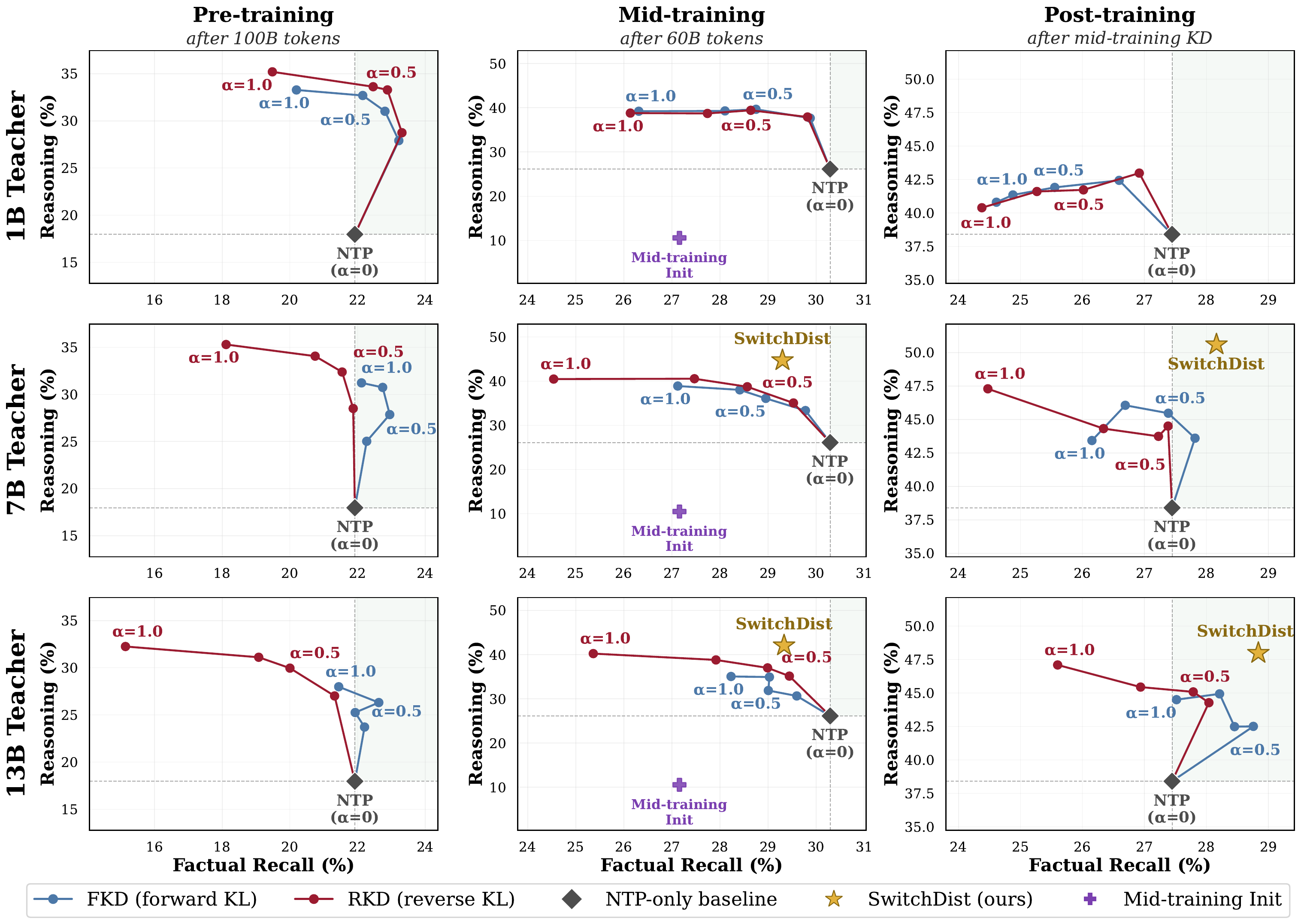}
   \vspace{-16pt}
    \caption{
    \textbf{Knowledge distillation exhibits a reasoning-recall tradeoff that falls below the NTP frontier during mid-training.} 
\textcolor{sdgold}{\ours{}} mitigates this tradeoff, and yields Pareto improvements over NTP after post-training. 
Rows correspond to 1B, 7B, and 13B teachers. We sweep over the distillation weight $\alpha$ for \textcolor{fklblue}{forward KL} and \textcolor{rklred}{reverse KL} distillation; \blackdiamond{} denotes the NTP baseline ($\alpha=0$), and mid-training panels show the \textcolor{studentpurple}{pre-trained student at initialization}.
    }
    \label{fig:frontier}
\end{figure*}

\cref{fig:frontier} compares reasoning and factual recall across distillation strength ($\alpha \in [0.0, 0.3, 0.5, 0.7, 1.0]$), KL direction (forward and reverse), training regime (pre-training and mid-training), and Instruct teacher size (1B, 7B, and 13B).\footnote{We observe the same qualitative trends for another model family, SmolLM2~\citep{allal2025smollm}, using a 1.7B Instruct teacher and 360M student, in \appref{app:smollm}.}

\paragraph{Increasing distillation strength consistently shifts performance toward reasoning.} Across teacher sizes and KL directions, increasing the contribution of the teacher (via larger $\alpha$) generally moves the operating point toward higher reasoning performance, with diminishing returns at stronger distillation. 
This trend holds across both pre-training and mid-training, suggesting that the teacher can reliably impart reasoning-relevant behavior even when the student has already acquired substantial knowledge.
This is consistent with recent work using knowledge distillation to transfer reasoning capabilities from stronger teachers~\citep{kim2026explain}.

\paragraph{The effect of distillation on factual recall is stage-dependent.} During pre-training (\cref{fig:frontier} left), moderate forward-KL distillation tends to improve factual recall alongside reasoning, yielding Pareto improvements over NTP across teacher sizes. At stronger distillation strengths, however, factual recall begins to decline especially under reverse KL. During mid-training (\cref{fig:frontier} middle), the pattern changes such that no KD operating point outperforms NTP on factual recall even as reasoning improves. Post-training changes this tradeoff asymmetrically: after applying the same standard post-training procedure to all mid-trained settings, without further distillation (\cref{fig:frontier} right), the factual-recall gap relative to NTP narrows or reverses for larger teachers, while the reasoning gains from distillation persist. This convergence partly reflects greater factual degradation of NTP during post-training, whereas distilled checkpoints retain more of their mid-training recall.

\paragraph{Thus, the reasoning–recall tradeoff changes qualitatively once distillation is applied to a substantially pre-trained student.} Because our pre-training and mid-training experiments use the same data mixture, this shift cannot be attributed to differences in training data. Instead, it points to an interaction between teacher supervision and the student’s prior training state: unlike a randomly initialized student, the mid-training student has already acquired substantial knowledge through pre-training. We investigate this interaction in the next section.

\begin{figure*}[t]
    \centering
    \includegraphics[width=\textwidth]{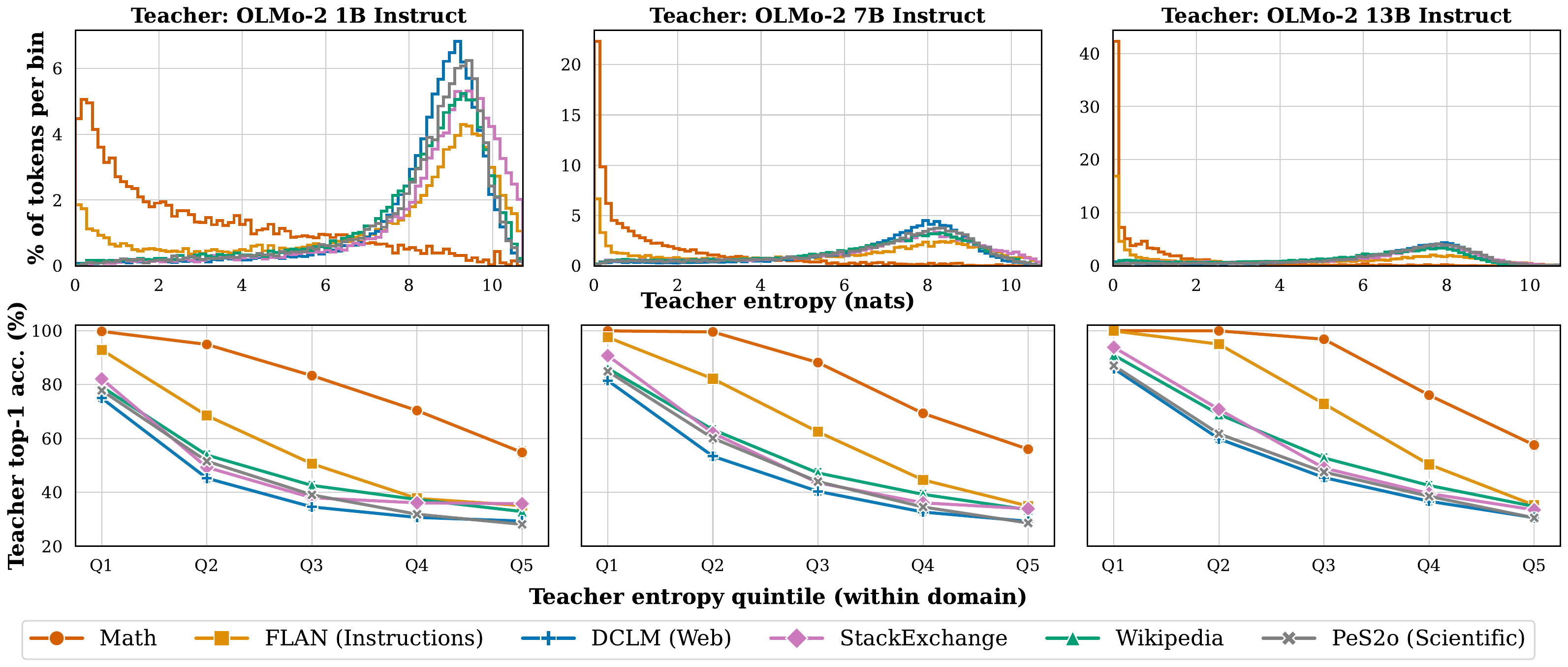}
   \vspace{-16pt}
    \caption{
\textbf{Low teacher entropy identifies tokens for which teacher supervision is most reliable.}
\textbf{Top:} Across teacher sizes, procedural domains (e.g., math, instruction-following) concentrate at lower teacher predictive entropy than unstructured domains.
\textbf{Bottom:} For each domain, lower entropy tokens exhibit substantially higher teacher top-1 agreement with the ground-truth token. Q1: lowest entropy; Q5: highest entropy.}
 \label{fig:teacher_supervision_asymmetric}
\end{figure*}

\section{Why Does the Reasoning--Recall Tradeoff Occur?}
\label{sec:analysis}
We attribute KD's stage-dependent behavior to the interaction of three factors:
(i) the teacher's predictive confidence,
(ii) the student's existing knowledge state, and (iii) the optimization dynamics induced by distillation. We study each factor in turn below.

\subsection{Teacher supervision is asymmetric across data domains}
\label{subsec:teacher_supervision}
We begin by asking whether teacher supervision is uniformly reliable across the training corpus. Using \olmotwo{} Instruct 1B, 7B, and 13B as teachers, we compute the \emph{teacher predictive entropy} $H_n$ for every token at position $n$ from randomly sampled Dolmino documents. Formally, let
\begin{align} \label{eq:entropy}
H_n &=- \sum_{v\in\mathcal V}
p_{\mathrm T}^{(\tau)}(v\mid x_{<n})
\log
p_{\mathrm T}^{(\tau)}(v\mid x_{<n})
\end{align}
denote the entropy of the teacher distribution $T$ (scaled with temperature $\tau$, and with vocabulary $\mathcal{V}$). Teacher entropy varies systematically across domains (\cref{fig:teacher_supervision_asymmetric}, top): procedural domains (e.g., math and instruction-following) exhibit substantially lower entropy than unstructured text domains.\footnote{This pattern generalizes across model families and training stages (\appref{app:more_analysis}).} Domain-level differences alone, however, do not establish that predictive entropy reflects supervision quality. Teacher predictive entropy is also indicative of correctness \emph{within} each domain, where correctness is defined as agreement with the ground-truth corpus token. In \cref{fig:teacher_supervision_asymmetric} (bottom), the probability that the teacher's top-1 prediction equals the ground-truth token decreases monotonically across entropy quintiles for every data domain and teacher size, showing that teacher entropy is predictive of supervision quality. 

Together, these results show that teacher supervision is asymmetric across domains: in distillation, the teacher provides confident, lower-entropy supervision on tokens from procedural domains, but substantially more diffuse, higher-entropy supervision on tokens from unstructured ones. This asymmetry may bias distillation toward learning reasoning better over factual recall. 

\subsection{Teacher entropy predicts student factual acquisition}
\label{subsec:student_side}
Given qualitatively consistent trends across teacher sizes, we use \olmotwo{} 7B Instruct as a representative teacher for the remaining analysis; corresponding 1B and 13B results are shown in  \appref{app:analyses_teacher_sizes}.

We track the acquisition of open-ended factual recall across NTP checkpoints on our \textsc{Factual Recall} tasks (TriviaQA, Natural Questions, and SimpleQA). For each question-answer sample, we teacher-force the prompt to the first answer token and measure (i) the teacher entropy at that position and (ii) whether the student’s top-1 prediction matches the first token of any gold answer alias. We use this token-level criterion to operationalize whether a fact has been acquired. Teacher entropy is used only to stratify factual examples; students are trained purely with NTP.

\begin{figure*}[t]
    \centering
    \includegraphics[width=\textwidth]{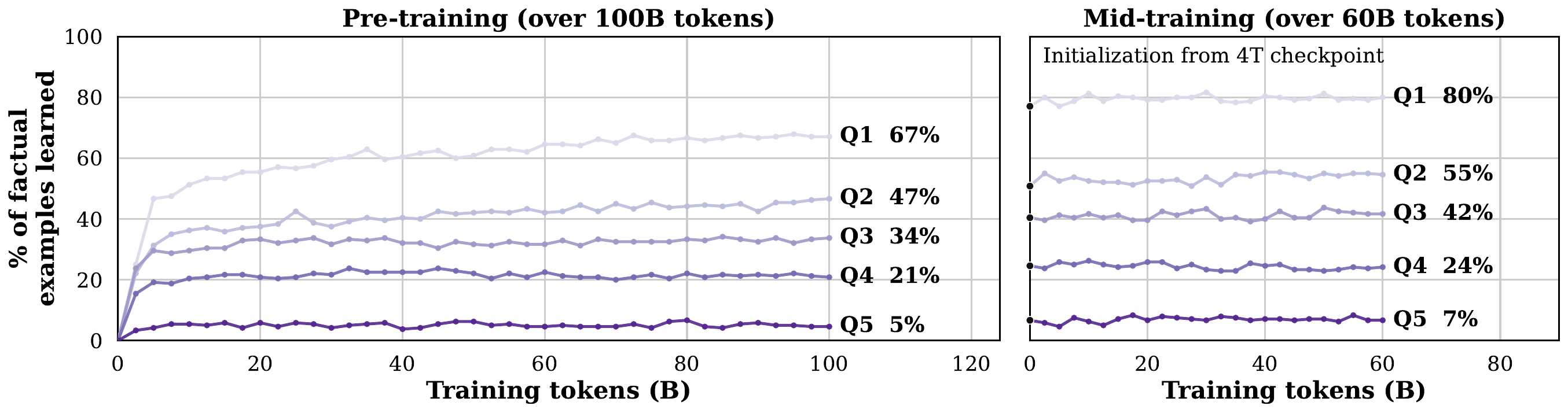}
    \vspace{-18pt}
\caption{
\textbf{Unresolved facts become concentrated at higher teacher entropy (lower quintiles).}
Lower-entropy facts are acquired earlier during pre-training; by mid-training initialization (after 4T tokens), unresolved facts are concentrated in the highest-entropy quintiles. Q1: lowest entropy; Q5: highest entropy. Results for other teacher sizes are in \cref{fig:quintile_acquisition_1b13b}.
}
    \label{fig:quintile_acquisition}
\end{figure*}

As shown in \cref{fig:quintile_acquisition}, teacher entropy strongly predicts factual acquisition: by the end of pre-training, the student has learned 67\% of facts in the lowest-entropy quintile (Q1), and only 5\% in the highest (Q5), with intermediate quintiles progressing monotonically. By the start of mid-training, this stratification has largely saturated, leaving unresolved factual-recall examples concentrated in the highest-entropy quintiles, which is precisely where teacher supervision is least confident.

\begin{figure*}[t]
    \centering
    \includegraphics[width=\textwidth]{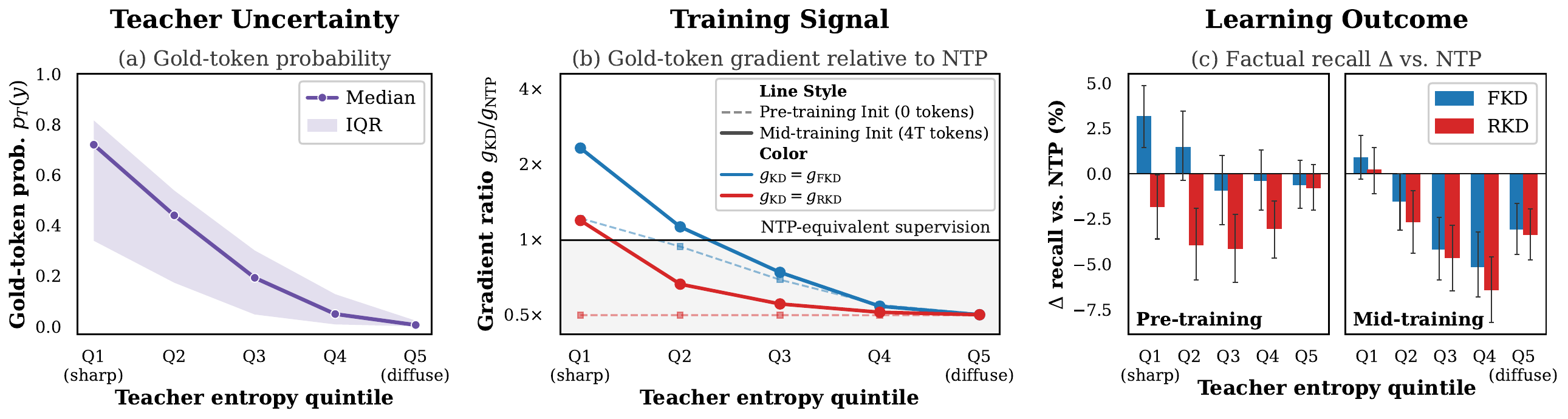}
 \vspace{-16pt}
    \caption{
\textbf{High teacher entropy provides weaker supervision for factual acquisition, reducing factual recall.}
Upon stratifying factual recall examples into teacher-entropy quintiles using an \olmotwo{} 7B Instruct teacher, the highest teacher entropy tokens (a) have lower teacher-assigned ground-truth probability, which (b) produces weaker optimization signal via gradient updates during KD, and (c) results in lower factual recall. Q1: lowest entropy; Q5: highest entropy. 
}
\label{fig:gold_attenuation}
\end{figure*}

\subsection{Knowledge distillation attenuates factual supervision}
\label{subsec:gradient_analysis}
Having characterized the teacher’s predictive confidence and the student’s knowledge state, we next examine their interaction through the distillation objective. Using the entropy-stratified factual recall examples from \cref{subsec:student_side}, we trace three cascading effects: (i) the teacher's probability assigned to that token, (ii) the supervision placed on the token under forward and reverse KD, and (iii) the resulting factual recall difference between KD and NTP. Across all three analyses, higher teacher entropy consistently corresponds to progressively weaker factual supervision and worse performance. 

First, teacher probability on the ground-truth token strictly decreases with predictive entropy (\cref{fig:gold_attenuation}a), meaning that the teacher places less weight on the correct answer as entropy increases. 

We next measure the gold-token directional gradient $g$, which captures how strongly the objective reinforces the correct next token. Because $g$ depends on both teacher and student distributions, we consider two student initializations representing the start of pre- and mid-training: random initialization and the 4T-token checkpoint, respectively. We compute the ratio of the gold-token gradient under FKD ($g_{\text{FKD}}$) or RKD ($g_{\text{RKD}}$) to NTP ($g_{\text{NTP}}$).\footnote{We evaluate at $\alpha=0.5$. Closed-form derivations are provided in \appref{app:gold_answer_gradient_derivation}.} Ratios below 1 indicate weaker ground-truth supervision than NTP. Both FKD and RKD increasingly attenuate ground-truth supervision as teacher entropy rises, reaching approximately $0.5\times$ NTP for the highest-entropy facts (\cref{fig:gold_attenuation}b).

To connect this weakened gradient signal to downstream acquisition, we compare KD and NTP factual recall within each entropy quintile (\cref{fig:gold_attenuation}c). After pre-training, FKD outperforms NTP on low-entropy facts (Q1--Q2), but this advantage disappears as entropy rises; RKD underperforms NTP across all quintiles at this stage. During mid-training, this effect is substantially stronger: low-entropy facts have largely already been acquired, leaving unresolved facts concentrated in high-entropy regions where KD provides the weakest supervision. Consequently, both FKD and RKD incur their largest factual recall deficits on high-entropy facts.

\section{\ours{} Improves the Reasoning--Recall Tradeoff}
\label{sec:method}
\begin{figure*}[t]
    \centering
    \includegraphics[width=\textwidth]{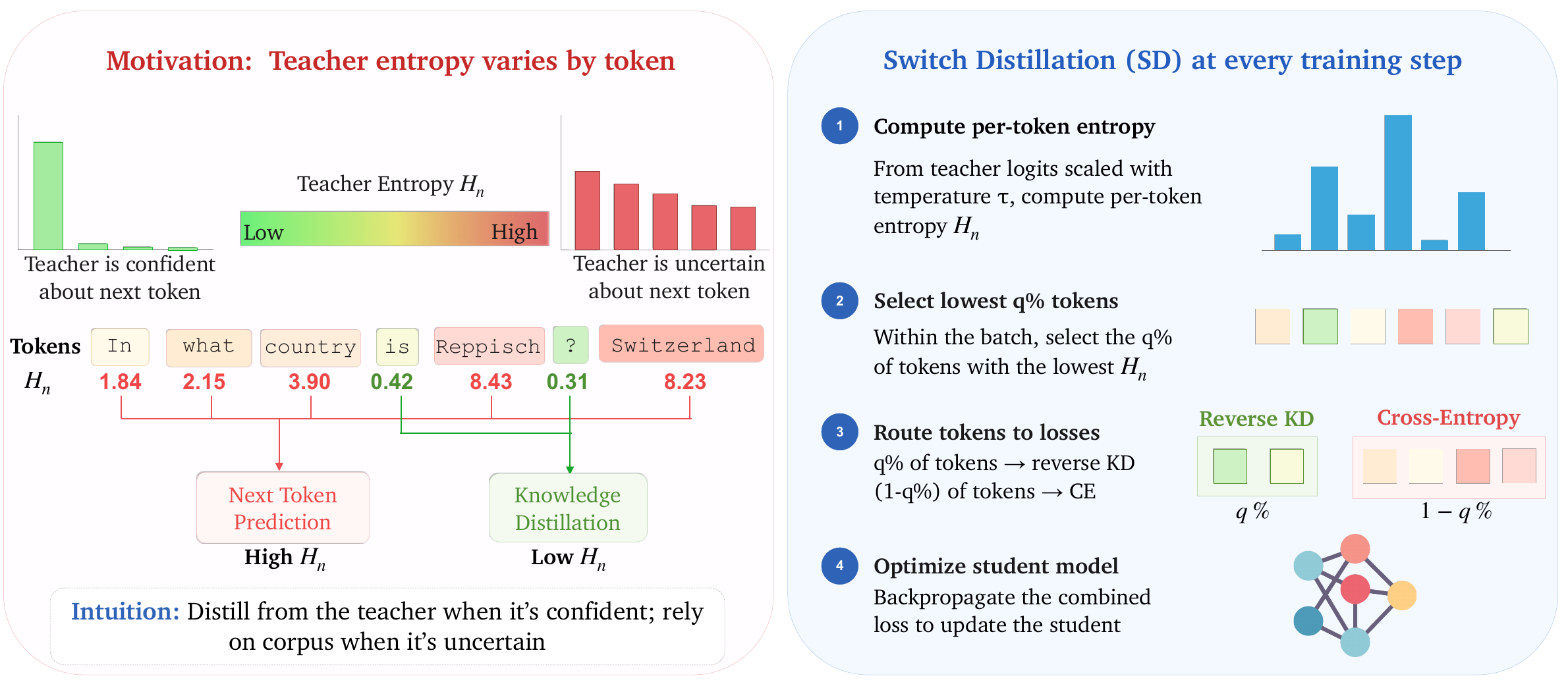}
\vspace{-16pt}
    \caption{
   \textbf{\ours{} overview.}}
    \label{fig:switchdist_method}
 \vspace{-16pt}
\end{figure*}
Evidence thus far suggests that teacher supervision is not equally beneficial across all tokens: teacher predictions tend to be concentrated on procedural reasoning trajectories but more diffuse on factual payload tokens. This suggests that a uniform distillation objective may be suboptimal during mid-training, and that teacher supervision should be applied only at token positions where it is most reliable. To this end, we introduce \ours{} (\cref{fig:switchdist_method}). 

Using teacher entropy $H_n$ (\cref{eq:entropy}), we route the lowest-$q\%$ of in-batch tokens to the reverse KL loss,\footnote{We normalize the terms separately so that each partition’s aggregate contribution is independent of its size; consequently, relative per-token weights vary with $q$. We select $q=20\%$; full sweep results are in \appref{app:q_tuning}.} defining 
$\mathcal S_q = \left\{ n: H_n \le \operatorname{Quantile}_q(\{H_n\}) \right\}$, the set of tokens assigned to distillation. 
Reverse KL is particularly well-suited to low-entropy teacher predictions as its mode-seeking behavior reinforces the teacher’s preferred continuation.
We optimize the following objective
\begin{align}
\mathcal L^{\textrm{\textsc{SwitchDist}}}&= \tau^2 \frac{1}{|\mathcal S_q|} \sum_{n\in\mathcal S_q} \mathrm{RKL} \!\left( p_{\mathrm S,n}^{(\tau)} \,\Vert\, p_{\mathrm T,n}^{(\tau)} \right)  + \frac{1}{|\bar{\mathcal S}_q|} \sum_{n\in \bar{\mathcal S}_q} \mathcal L_{\mathrm{CE},n}, \label{eq:switch_distillation}
\end{align}
where $\bar{\mathcal S}_q$ denotes the complement of $\mathcal S_q$ over supervised tokens. Note that \ours{} adds only negligible entropy and quantile computations beyond standard online KD, requiring no additional parameters or model forward passes. 

\section{\ours{} Experiments}
\subsection{Experimental setup} \label{subsec:exp_setup}
We evaluate \ours{} under the same mid-training setup as in \cref{subsec:experimental}, and consider \olmotwo{} 7B Instruct and 13B Instruct as teacher models. Our baselines include standard next-token prediction (\textbf{NTP}), as well as forward and reverse knowledge distillation (\textbf{FKD} and \textbf{RKD}, respectively) at $\alpha=0.5$, which empirically provides the best balance between factual recall and reasoning. 
We also compare against token-routing KD (\textbf{TRKD})~\citep{goyal2026distilled}.
TRKD applies forward-KL distillation to high-entropy tokens while retaining CE on all tokens, whereas \ours{} hard-switches between reverse-KL on low-entropy tokens and CE otherwise.\footnote{TRKD was originally proposed to improve in-context learning; we provide a more detailed description of the method and its differences from \ours{} in \cref{sec:related_work}.}

\subsection{\ours{} improves reasoning while preserving factual recall} \label{subsec:midtrain_results}
\cref{tab:mid_training_results} compares \ours{} against mid-training baselines. 
Across both teacher sizes, \ours{} achieves the strongest \textsc{Reasoning}, improving the macro-average from 26.1\% under NTP to 44.7\%/42.1\% with the 7B and 13B teachers, while remaining the KD baseline closest to NTP on \textsc{Factual Recall} (29.3\%/29.3\% vs.\ 30.3\%). \ours{} also achieves the strongest \textsc{Knowledge \& Commonsense} performance (49.3\%/46.5\%). Distillation with the 7B teacher generally outperforms the 13B teacher, corroborating work showing that a large size difference between teacher and student---defined as the \emph{capacity gap}---may reduce distillation effectiveness~\citep{mirzadeh2019improvedknowledgedistillationteacher, panigrahi2024progressivedistillationinducesimplicit, busbridge2025distillation}.

\begin{table*}[t]
\centering
\tiny
\setlength{\tabcolsep}{2.8pt}
\renewcommand{\arraystretch}{1.15}
\caption{
\textbf{Full downstream results after mid-training.}
The NTP baseline is duplicated because it has no teacher ($T=\mathrm{N/A}$) and therefore serves as the shared reference for both teacher size blocks.
\textbf{Bold} denotes the best result per teacher block, and $^{*}$ indicates a statistically significant improvement over the strongest competing baseline ($p<0.05$, paired bootstrap).
Benchmark names are abbreviated; see \cref{tab:evaluation_tasks} for full task names.
}
\vspace{-10pt}
\setlength{\tabcolsep}{1pt}
\resizebox{\columnwidth}{!}{%
\begin{tabular}{ll|cccccc|ccc|cccccc}
\toprule
&
&
\multicolumn{6}{c|}{Reasoning}
&
\multicolumn{3}{c|}{Factual Recall}
&
\multicolumn{6}{c}{Knowledge \& Commonsense} \\
\cmidrule(lr){3-8}
\cmidrule(lr){9-11}
\cmidrule(l){12-17}
$T$ & Method
& GSM8K & GSM-S & GSM+ & BBH & DROP & MATH
& TQA & NQ & SQA
& MMLU & MMLU-P & ARC-C & OBQA & Wino & AGI \\
\midrule

N/A & NTP & 40.4 & 29.8 & 23.1 & 29.9 & 29.8 & 3.8 & \textbf{56.7}$^{*}$ & \textbf{25.5} & \textbf{8.7} & 43.6 & 15.5 & 51.1 & 51.8 & 51.4 & 34.1 \\

7B & FKD & 57.8 & 42.6 & 36.2 & 30.6 & 41.9 & 7.6 & 53.9 & 24.7 & 8.4 & 49.5 & 18.7 & 61.9 & 61.8 & 51.5 & 38.6 \\

7B & RKD & 62.1 & 46.2 & 39.2 & 31.6 & 43.1 & 10.4 & 53.1 & 24.1 & 8.4 & 50.1 & 18.9 & 62.5 & 62.8 & 52.5 & 39.3 \\

7B & TRKD & 52.5 & 36.9 & 31.0 & 30.3 & 36.8 & 5.6 & 54.0 & 24.1 & 7.8 & 47.9 & 17.4 & 58.9 & 60.0 & 51.2 & 37.5 \\

7B & \textbf{SD} & \textbf{69.7}$^{*}$ & \textbf{55.3}$^{*}$ & \textbf{46.1}$^{*}$ & \textbf{32.8}$^{*}$ & \textbf{49.6}$^{*}$ & \textbf{14.8}$^{*}$ & 54.9 & 24.6 & 8.4 & \textbf{51.6}$^{*}$ & \textbf{19.8}$^{*}$ & \textbf{64.7}$^{*}$ & \textbf{64.2} & \textbf{53.8} & \textbf{41.5}$^{*}$ \\

\midrule[0.8pt]

N/A & NTP & 40.4 & 29.8 & 23.1 & 29.9 & 29.8 & 3.8 & \textbf{56.7}$^{*}$ & \textbf{25.5} & 8.7 & 43.6 & 15.5 & 51.1 & 51.8 & 51.4 & 34.1 \\

13B & FKD & 52.7 & 37.9 & 31.8 & 29.4 & 33.1 & 6.4 & 54.2 & 24.7 & 8.1 & 47.5 & 16.3 & 55.3 & 56.2 & 51.2 & 36.2 \\

13B & RKD & 59.2 & 48.7 & 37.4 & 31.4 & 37.3 & 8.0 & 53.4 & 24.0 & 8.5 & 48.4 & 17.3 & 57.8 & 60.0 & 51.0 & 38.3 \\

13B & TRKD & 47.8 & 32.7 & 28.4 & 29.3 & 34.0 & 5.6 & 53.8 & 24.2 & 8.3 & 45.7 & 15.4 & 53.0 & 56.6 & 50.7 & 35.2 \\

13B & \textbf{SD} & \textbf{66.0}$^{*}$ & \textbf{54.8}$^{*}$ & \textbf{42.2}$^{*}$ & \textbf{31.6} & \textbf{45.8}$^{*}$ & \textbf{12.0}$^{*}$ & 54.1 & 24.9 & \textbf{9.0} & \textbf{48.5} & \textbf{17.7} & \textbf{58.5} & \textbf{63.8}$^{*}$ & \textbf{51.5} & \textbf{39.2} \\

\bottomrule
\end{tabular}}
\label{tab:mid_training_results}
\end{table*}

\subsection{\ours{}'s benefits persist through post-training} \label{subsec:posttrain_results}
We next ask whether \ours{}'s mid-training gains are sustained after post-training, as stronger base models do not necessarily translate to stronger final models after alignment~\citep{springer2025overtrained, lu2026strongteacherneededdistillation, watts2026sharpnessawarepretrainingmitigatescatastrophic}. We apply \olmotwo{} 1B's four-stage post-training pipeline---supervised fine-tuning (SFT), direct preference optimization (DPO), and two rounds of reinforcement learning with verifiable rewards (RLVR1, RLVR2)---to each mid-trained model and report final performance in \cref{tab:post_training_rlvr2_results}.\footnote{We report full evaluation results after each intermediate post-training stage in \appref{app:full_results}.} Post-training improves reasoning across all methods, and \ours{} remains the strongest method on the \textsc{Reasoning} task group, with significant gains on all six tasks with the 7B teacher and five of six with the 13B teacher. With the 7B teacher, \ours{}’s macro-average increases from 44.7\% to 50.6\%; with the 13B teacher, it increases from 42.1\% to 48.0\%. Consistent with work showing that post-training can degrade factual recall and broader knowledge~\citep{gekhman-etal-2024-fine, ghosal2024understanding,yuan2024holisticevaluationllmsfactual, kaplan2026finetuningencourageshallucinationsfix}, we observe modest degradation in \textsc{Knowledge \& Commonsense} and \textsc{Factual Recall} across methods. However, \ours{} is the most robust: despite entering post-training with a small factual-recall deficit relative to NTP, it experiences the least forgetting and finishes with the highest \textsc{Factual Recall} macro-average.

\begin{table*}[t]
\centering
\tiny
\setlength{\tabcolsep}{2.8pt}
\renewcommand{\arraystretch}{1.15}
\caption{
\textbf{Downstream results after post-training.}
Notation and significance testing follow \cref{tab:mid_training_results}.
}
\vspace{-10pt}
\setlength{\tabcolsep}{1pt}
\resizebox{\columnwidth}{!}{%
\begin{tabular}{ll|cccccc|ccc|cccccc|c}
\toprule
&
&
\multicolumn{6}{c|}{Reasoning}
&
\multicolumn{3}{c|}{Factual Recall}
&
\multicolumn{6}{c|}{Knowledge \& Commonsense}
&
\multicolumn{1}{c}{Inst.} \\
\cmidrule(lr){3-8}
\cmidrule(lr){9-11}
\cmidrule(lr){12-17}
\cmidrule(l){18-18}
$T$ & Method
& GSM8K & GSM-S & GSM+ & BBH & DROP & MATH
& TQA & NQ & SQA
& MMLU & MMLU-P & ARC-C & OBQA & Wino & AGI
& IFE \\
\midrule

N/A & NTP & 67.0 & 46.8 & 40.6 & 31.1 & 32.7 & 12.6 & 53.3 & 21.6 & 7.5 & 32.3 & 16.6 & 52.4 & 52.4 & 51.4 & 32.3 & 62.1 \\

7B & FKD & 76.2 & 56.4 & 51.6 & 33.8 & 38.4 & 18.0 & 51.9 & 22.2 & 8.1 & 42.2 & 18.6 & 60.2 & 57.4 & 51.5 & 38.0 & 64.7 \\

7B & RKD & 73.8 & 53.4 & 49.5 & 31.6 & 40.5 & 17.6 & 51.8 & 21.9 & 7.9 & 46.2 & 18.0 & 61.1 & 53.0 & 51.2 & 37.1 & 61.6 \\

7B & TRKD & 70.7 & 52.9 & 46.3 & 31.8 & 36.8 & 14.4 & 51.7 & 22.3 & 8.0 & 40.9 & 17.4 & 59.0 & 53.6 & 51.5 & 35.6 & 65.2 \\

7B & \textbf{SD} & \textbf{79.8}$^{*}$ & \textbf{65.0}$^{*}$ & \textbf{52.7}$^{*}$ & \textbf{35.6}$^{*}$ & \textbf{48.2}$^{*}$ & \textbf{22.4}$^{*}$ & \textbf{53.6} & \textbf{22.7} & \textbf{8.2} & \textbf{48.1}$^{*}$ & \textbf{19.8}$^{*}$ & \textbf{62.6} & \textbf{59.4} & \textbf{54.6}$^{*}$ & \textbf{39.6}$^{*}$ & \textbf{69.5}$^{*}$ \\

\midrule[0.8pt]

N/A & NTP & 67.0 & 46.8 & 40.6 & 31.1 & 32.7 & 12.6 & 53.3 & 21.6 & 7.5 & 32.3 & 16.6 & 52.4 & 52.4 & 51.4 & 32.3 & 62.1 \\

13B & FKD & 72.1 & 54.3 & 46.5 & 31.4 & 35.7 & 14.6 & 54.1 & 23.2 & 8.0 & 36.1 & \textbf{18.1} & 55.0 & 56.4 & 51.2 & 35.9 & 64.9 \\

13B & RKD & 75.1 & 58.1 & 49.3 & \textbf{33.3} & 38.7 & 16.0 & 53.9 & 22.3 & 7.7 & 38.2 & 17.0 & 56.0 & 55.2 & 51.1 & 36.6 & 64.1 \\

13B & TRKD & 69.6 & 48.5 & 42.6 & 31.5 & 34.6 & 12.2 & 53.9 & 22.2 & 8.2 & 37.0 & 17.3 & 55.4 & 54.0 & 51.3 & 33.6 & 64.1 \\

13B & \textbf{SD} & \textbf{77.8}$^{*}$ & \textbf{62.8}$^{*}$ & \textbf{51.8}$^{*}$ & \textbf{33.3} & \textbf{42.8}$^{*}$ & \textbf{19.6}$^{*}$ & \textbf{54.2} & \textbf{24.1} & \textbf{8.4} & \textbf{43.1}$^{*}$ & \textbf{18.1} & \textbf{56.5} & \textbf{59.0} & \textbf{52.5} & \textbf{38.1} & \textbf{67.1} \\

\bottomrule
\end{tabular}}
\label{tab:post_training_rlvr2_results}
\end{table*}

\subsection{Ablations} 
\label{subsec:ablations}
We further ablate the design choices to \ours{} using the \olmotwo{} 7B Instruct teacher and report mid-training macro-averages in \cref{tab:ablation_results} (per-task results in \appref{app:ablation_full_results}).

\noindent
\begin{minipage}[t]{0.60\linewidth}
\vspace{0pt}
\centering
\captionof{table}{\textbf{Ablation results.} The first row reports \ours{}'s absolute performance; subsequent rows report relative changes from other design choices.}
\label{tab:ablation_results}
\vspace{-7pt}
\centering
\scriptsize
\resizebox{\linewidth}{!}{%
\begin{tabular}{lccc}
\toprule
\textbf{Method / Ablation}
    & \textbf{Reasoning}
    & \textbf{Factual Recall}
    & \textbf{Knowledge} \\
\midrule

\ours{} 
    & \cellcolor{gray!15}44.7
    & \cellcolor{gray!15}29.3
    & \cellcolor{gray!15}49.3 \\

\midrule
\multicolumn{4}{l}{\textbf{Distillation Objective}} \\
\cmidrule(lr){1-4}

\ours{}$_{\textsc{FKL}}$
    & \cellcolor{red!26}-2.9
    & \cellcolor{red!2}-0.2
    & \cellcolor{red!13}-1.3 \\

\midrule
\multicolumn{4}{l}{\textbf{Routing Policy}} \\
\cmidrule(lr){1-4}

Teacher-Correct Routing
    & \cellcolor{red!40}-4.4
    & \cellcolor{red!46}-5.1
    & \cellcolor{red!9}-1.0 \\
Random Routing
    & \cellcolor{red!59}-6.5
    & \cellcolor{red!7}-0.8
    & \cellcolor{red!18}-2.0 \\
Oracle Domain Routing
    & \cellcolor{red!65}-7.2
    & \cellcolor{red!12}-1.3
    & \cellcolor{red!21}-2.3 \\
\midrule
\multicolumn{4}{l}{\textbf{Supervision Objective}} \\
\cmidrule(lr){1-4}

Always CE
    & \cellcolor{red!3}-0.3
    & \cellcolor{gray!5}+0.1
    & \cellcolor{red!5}-0.6 \\
Teacher Top-1 Labels
    & \cellcolor{red!58}-6.4
    & \cellcolor{green!12}+1.3
    & \cellcolor{red!25}-2.8 \\

\bottomrule
\end{tabular}
}

\end{minipage}%
\hfill
\begin{minipage}[t]{0.38\linewidth}
\vspace{0pt}

We examine three design choices using the \olmotwo{} 7B Instruct teacher: \emph{KL direction}, replacing RKL with FKL (\textbf{\ours{}$_{\mathrm{FKL}}$}); \emph{routing signal}, replacing teacher entropy with whether the teacher's top-1 prediction matches the target (\textbf{Teacher-Correct Routing}) or with a random mask (\textbf{Random Routing}) (each with a fixed routing budget of $q=20\%$), or whether the token comes from MATH or FLAN (\textbf{Oracle Domain Routing}); and \emph{supervision objective}, retaining CE on all tokens (\textbf{Always CE}) or replacing soft distillation with the teacher's top-1 predictions (\textbf{Teacher Top-1 Labels}).
\end{minipage}

Replacing RKL with FKL modestly reduces \textsc{Reasoning} and \textsc{Knowledge \& Commonsense}, suggesting that RKL may better exploit the sharp, low-entropy teacher distributions selected by our routing strategy. Alternative routing signals consistently underperform teacher entropy, while \textbf{Always CE} has little effect. Finally, \textbf{Teacher Top-1 Labels} improves \textsc{Factual Recall} at the expense of the other categories. These results identify entropy-based routing as the driver of our method's gains, while soft teacher distributions provide useful supervision beyond top-1 predictions.

\subsection{Related Work}
\label{sec:related_work}
\paragraph{Mid-training and its origins.}
Modern foundation model development has converged on mid-training as a distinct stage between pre-training and post-training, during which models are further optimized in a self-supervised fashion on curated data mixtures~\citep{zhang2025interplaypretrainingmidtrainingrl, liu2026midtrainingbridgespretrainingposttraining}. While mid-training has its roots in continued pre-training, its modern formulation emphasizes capability-focused data mixtures designed to better prime models for subsequent post-training~\citep{gururangan-etal-2020-dont}. 

Recent work has indicated that surfacing post-training capabilities earlier during mid-training can further strengthen desirable downstream attributes; carefully designed mid-training recipes have been found to particularly benefit subsequent reinforcement learning~\citep{wang2025octothinkermidtrainingincentivizesreinforcement, huang2026remitrlguidedmidtrainingiterative, liu2026midtrainingbridgespretrainingposttraining, tan2026selfimprovingpretrainingusingposttrained}. 
To date, most of the mid-training literature employs the standard next-token prediction objective. We revisit knowledge distillation in this setting and uncover a previously uncharacterized reasoning--recall tradeoff. 

\paragraph{Data-efficient language modeling.} As the growth of compute is on track to outpace the supply of organic web text, high-quality human-written data is becoming increasingly scarce for language model training~\citep{kim2026pretraining}. This impending constraint has motivated a body of work on \emph{data-efficient language modeling}.
Prior approaches have largely pursued data efficiency by improving the training corpus itself through curation, augmentation, selection, or mixture optimization~\citep{gunasekar2023textbooksneed, xie2023data,xie2023doremi,lin2024not,maini-etal-2024-rephrasing, nguyen2025recycling,chen2026olmix, kim2026dataefficientpretrainingscalingsynthetic}, often with guidance from stronger reference models. More broadly, \citet{kim2026pretraining} argue that algorithmic interventions such as ensembling or self-distillation may soon serve as important avenues for tackling the data wall. We study this algorithmic perspective at the mid-training stage, which necessitates substantially higher-quality data than large-scale pre-training while consuming orders of magnitude more tokens than post-training. Our approach is complementary to these data-centric methods: rather than modifying the training corpus, we improve how supervision is extracted from each observed token.

\paragraph{Knowledge distillation across training stages.}
Knowledge distillation broadly encompasses both hard (sequence-level) distillation, in which a teacher generates synthetic training sequences for subsequent next-token prediction~\citep{kim2016sequencelevelknowledgedistillation}, and soft (logit-based) distillation, in which the student is trained to match the teacher's predictions by minimizing its distributional divergence~\citep{hinton2015distillingknowledgeneuralnetwork}. We focus on the latter, whose role has been studied extensively during language model pre-training and post-training. 

In the pre-training setting, \citet{busbridge2025distillation} derive scaling laws for language model distillation and characterize compute-optimal teacher-student configurations under fixed compute budgets. 
\citet{goyal2026distilled} show that pre-training distillation improves test-time scaling at the cost of in-context learning and propose token-routing KD (TRKD) to mitigate this degradation. While TRKD targets in-context learning during pre-training, \ours{} addresses the reasoning--recall tradeoff that emerges during mid-training.
Finally, \citet{cha2026why} show that generative distillation induces a precision–recall tradeoff, whereby lower-entropy teachers produce sharper but lower-coverage students.

Recent work has revisited the standard KD formula primarily in the post-training setting. MiniLLM, for one, advocates reverse-KL distillation to improve generative capabilities of LMs, while subsequent work proposes adaptive combinations of forward and reverse KL to exploit their distinct optimization behaviors~\citep{gu2024minillm,zhong-etal-2024-revisiting,wu-etal-2025-rethinking}. Reverse-KL is the standard KD direction for \emph{on-policy knowledge distillation}, a relatively new post-training paradigm in which the student is trained on its own sampled trajectories under teacher guidance~\citep{agarwal2024onpolicydistillationlanguagemodels}. Token-Selective Dual Knowledge Distillation (TSD-KD)~\citep{kim2026explain} is an on-policy KD method that selectively applies teacher supervision based on teacher–student confidence discrepancies and teacher-ranked student-generated reasoning trajectories. Entropy-Aware On-Policy Distillation (EOPD) proposes to use teacher predictive entropy as a signal to interpolate between reverse and forward KL for high-entropy teacher distributions~\citep{jin2026entropyawareonpolicydistillationlanguage}. Among prior work, EOPD is conceptually closest to \ours{}, but differs in both training regime and use of entropy: EOPD operates during post-training on student-generated trajectories and always applies teacher-based distillation, adapting the KL objective as teacher uncertainty varies. In contrast, \ours{} operates at the self-supervised mid-training stage on fixed tokens and uses entropy to determine whether to obtain supervision from the corpus or the teacher model. 

Collectively, these works establish knowledge distillation as an effective supervision strategy during pre-training and post-training. We extend this line of work to the emerging mid-training regime and show that KD exhibits a fundamentally different tradeoff. This behavior consequently motivates a stage-specific adaptation of the standard distillation objective.

\section{Discussion}
Enabling language models to learn more from a fixed data pool remains a longstanding challenge, typically addressed by improving training data quality. We identify a complementary direction: improving \emph{how} existing tokens are learned during mid-training,  where high-quality tokens are scarce and expensive. We argue that knowledge distillation should not be stage-agnostic: while the standard formulation is effective during pre-training, it exhibits fundamentally different behavior during mid-training, where teacher uncertainty and student knowledge interact to produce a reasoning--recall tradeoff. By explicitly accounting for this interaction, \ours{} consistently improves reasoning while largely preserving factual recall in a token-matched setting. More broadly, our findings suggest that objectives themselves ought to be stage-aware. While we study mid-training, this principle may extend to other phases where the student’s knowledge has substantially evolved, such as late-stage or continual pre-training. We hope our findings motivate further investigation into stage-aware optimization methods for data-efficient language modeling.

\section*{Acknowledgments}
We thank (in alphabetical order) Millicent Li, Emmy Liu, Jacob Mitchell Springer, and Ishaan Watts for helpful technical discussions about this project, and Hamish Ivison for discussions about \olmotwo{} training. JH and SSL are supported by the Meta AI Mentorship Program; JH is additionally supported by an NSF Graduate Research Fellowship. PWK was supported by the Singapore National Research Foundation and the National AI Group in the Singapore Ministry of Digital Development and Information under the AI Visiting Professorship Programme (award number AIVP-2024-001) and the AI2050 program at Schmidt Sciences.

\clearpage
\newpage
\bibliographystyle{assets/plainnat}
\bibliography{paper}

\clearpage
\newpage
\section*{Appendix}
\startcontents[sections]  
\printcontents[sections]{l}{1}{\setcounter{tocdepth}{2}}
\newpage
\appendix
\section{General}
\subsection{Limitations}
\label{sec:limitations}
Our main experiments are conducted with the \olmotwo{} recipe, which enables controlled and replicable experimentation across pre-training, mid-training, and post-training. We provide supplementary evidence that our findings extend beyond this setting: the stage-dependent tradeoff exhibited by KD also appears in the SmolLM2 family (\appref{app:smollm}), while similar asymmetries in teacher supervision emerge across teachers from different training stages and model families (\appref{app:teacher_supervision_asymmetry}). Broader controlled validation is difficult because isolating stage-dependent distillation effects requires access to more than model weights: it requires intermediate pre- and mid-training checkpoints, the corresponding data mixtures, and sufficiently complete training recipes to reproduce transitions between stages. Few model families currently release all of these artifacts. Moreover, logit-based distillation mathematically requires compatible output vocabulary between student and teacher, further restricting the set of viable model pairs. We therefore center our controlled experiments on \olmotwo{}, with SmolLM2 and cross-family teacher analyses as complementary tests for generality. 

Similarly, we adopt competitive defaults wherever possible: while our student model sizes are relatively small (1B parameters), we train substantially beyond Chinchilla-optimal token budgets and distill from Instruct models that outperform their base counterparts both as standalone models and as teachers. We do not exhaustively ablate these experimental choices, such as the effect of teacher post-training or broader teacher scales, as doing so would require substantial additional compute and prior work has already characterized several of these dimensions; for example, increasing the teacher-student capacity gap can impair distillation effectiveness~\citep{mirzadeh2019improvedknowledgedistillationteacher}. Our experiments instead focus compute on isolating how distillation behavior changes across training stages and objectives.

Exploring better strategies for selective distillation is an interesting future extension. \ours{} routes supervision using teacher predictive entropy, a simple primitive that requires no additional supervision or parameters. While our proposed algorithm is cheap and effective, richer and more expressive strategies (i.e., a learned router network, analogous to those employed by Mixture-of-Experts architectures~\citep{shazeer2017outrageouslylargeneuralnetworks, li2026slicingdicingconfiguringoptimal}) may better capture when and where teacher supervision is beneficial. As \ours{} arises from our study on how best to leverage teacher supervision conditional on fixed data, it may be broadly compatible with approaches that instead optimize the data pool itself.  

Finally, while we develop and evaluate \ours{} as a \emph{mid-training} strategy, our tradeoff analysis suggests that it may be beneficial more broadly whenever factual acquisition slows under teacher supervision. In particular, we hypothesize that \ours{} may also improve upon standard KD during late-stage pre-training, when the student has already acquired much of the easily transferred knowledge from the teacher. Characterizing when standard KD ceases to yield Pareto improvements and begins to induce such a tradeoff is an important future direction.

\subsection{AI Usage Statement}
We used generative AI tools in this work for lightweight assistance with copy-editing the manuscript, creating and improving the presentation of scientific figures, debugging implementations, and automating our training and evaluation scripts. We did not use any generative AI for methodological or experimental design, the analysis and interpretation of results, or the identification of relevant prior work. All AI-assisted code, figures, and text were manually reviewed before use. We take full responsibility for the final content of this work, including text, claims or artifacts produced with the assistance of generative AI.

\subsection{Reproducibility Statement}
All our experiments are conducted using open-source training and evaluation stacks, with publicly available models and datasets. We provide complete training configuration, hyperparameter, and evaluation details in \appref{app:exp_details}. Moreover, \ours{} requires only a minimal modification to standard knowledge distillation; we describe the mechanics of it in considerable detail in \cref{sec:method}, and provide the pseudocode in \appref{app:pseudocode}.  

\section{\ours{} Details}

\subsection{Pseudocode}
\label{app:pseudocode}
We provide the pseudocode for \ours{} in \cref{alg:entropy_gated_rkl}.
\begin{algorithm}[t]
\caption{\ours{}}
\label{alg:entropy_gated_rkl}
\begin{algorithmic}[1]
\Require Student model $p_S$, teacher model $p_T$, local token batch $x_{1:B}$, routing quantile $q \in (0,1)$, temperature $\tau$
\Ensure Training loss $\mathcal L_{\ours}$

\State $\mathcal{B}_{\mathrm{tok}} \gets$ all valid next-token positions in $x_{1:B}$, with target $y_n$ at each position $n$

\State $z_T, z_S \gets \operatorname{Forward}(p_T, x_{1:B}), \operatorname{Forward}(p_S, x_{1:B})$
\Comment{Teacher and student logits}

\Statex
\Comment{Compute softened distributions used for logit-based distillation.}
\State $p_{T,n}^{(\tau)} \gets \operatorname{softmax}(z_T^n/\tau)$ for all $n\in\mathcal{B}_{\mathrm{tok}}$
\State $p_{S,n}^{(\tau)} \gets \operatorname{softmax}(z_S^n/\tau)$ for all $n\in\mathcal{B}_{\mathrm{tok}}$

\Statex
\Comment{Score tokens by teacher predictive entropy and route by quantile.}
\State $H_n \gets -\sum_{v\in\mathcal V} p_{T,n}^{(\tau)}(v)\log p_{T,n}^{(\tau)}(v)$ for all $n\in\mathcal{B}_{\mathrm{tok}}$
\State $\mathcal S_q \gets \left\{n\in\mathcal B_{\mathrm{tok}} :
H_n \le \operatorname{Quantile}_q\!\left(
\{H_{n'}:n'\in\mathcal B_{\mathrm{tok}}\}
\right)\right\}$
\Comment{Route low-entropy tokens to KD}
\State $\bar {\mathcal S_q} \gets \mathcal{B}_{\mathrm{tok}}\setminus \mathcal S_q$

\Statex
\Comment{Compute separately normalized RKL and CE objectives.}
\State
\[
\mathcal L_{\mathrm{RKL}}
\gets
\frac{\tau^2}{|\mathcal S_q|}
\sum_{n\in \mathcal S_q}
\mathrm{KL}\!\left(
p_{S,n}^{(\tau)}
\,\middle\|\,
p_{T,n}^{(\tau)}
\right)
\]

\State
\[
\mathcal L_{\mathrm{CE}}
\gets
\frac{1}{|\bar {\mathcal S_q}|}
\sum_{n\in\bar {\mathcal S_q}}
\left[-\log p_S(y_n\mid x_{<n})\right]
\]

\State $\mathcal L_{\ours}\gets\mathcal L_{\mathrm{RKL}}+\mathcal L_{\mathrm{CE}}$

\State \Return $\mathcal L_{\ours}$

\end{algorithmic}
\end{algorithm}

\subsection{Ablating $q$}
\label{app:q_tuning}
We sweep the routing threshold $q \in \{10\%, 20\%, 30\%\}$ for both the 7B and 13B teacher settings. \cref{tab:quantile_ablation_results} shows per-task results; overall, we choose $q=20\%$ as the default routing threshold across teacher sizes, as it provides the highest reasoning performance, while remaining competitive or best on factual recall and knowledge \& commonsense. Performance is relatively stable between $q=20\%$ and $q=30\%$, suggesting that the method is not highly sensitive to the precise routing threshold.
\begin{table*}[t]
\centering
\renewcommand{\arraystretch}{1.15}
\caption{
\textbf{Downstream results for \ours{} at routing thresholds $q \in \{10\%, 20\%, 30\%\}$.}
\textbf{Bold} denotes the best result per teacher block.
}
\setlength{\tabcolsep}{1pt}
\resizebox{\columnwidth}{!}{%
\begin{tabular}{ll|cccccc|c|ccc|c|cccccc|c}
\toprule
&
&
\multicolumn{7}{c|}{Reasoning}
&
\multicolumn{4}{c|}{Factual Recall}
&
\multicolumn{7}{c}{Knowledge \& Commonsense} \\
\cmidrule(lr){3-9}
\cmidrule(lr){10-13}
\cmidrule(l){14-20}
$T$ & Method
& GSM8K & GSM-S & GSM+ & BBH & DROP & MATH & \textbf{Avg.}
& TQA & NQ & SQA & \textbf{Avg.}
& MMLU & MMLU-P & ARC-C & OBQA & Wino & AGI & \textbf{Avg.} \\
\midrule

7B & $q=10\%$ & 61.6 & 48.8 & 39.4 & 32.5 & \textbf{50.2} & 11.4 & 40.6 & 53.7 & 23.9 & 8.6 & 28.7 & 50.6 & 19.0 & 62.0 & 63.8 & \textbf{55.5} & 40.4 & 48.5 \\

7B & $q=20\%$ & 69.7 & \textbf{55.3} & \textbf{46.1} & 32.8 & 49.6 & \textbf{14.8} & \textbf{44.7} & \textbf{54.9} & 24.6 & 8.4 & 29.3 & \textbf{51.6} & \textbf{19.8} & \textbf{64.7} & \textbf{64.2} & 53.8 & \textbf{41.5} & \textbf{49.3} \\

7B & $q=30\%$ & \textbf{69.8} & 54.1 & 45.6 & \textbf{33.8} & 47.2 & 13.0 & 43.9 & 54.7 & \textbf{25.3} & \textbf{8.8} & \textbf{29.6} & 51.0 & 19.5 & 63.5 & 62.4 & 53.0 & 40.7 & 48.4 \\

\midrule[0.8pt]

 13B & $q=10\%$ & 62.1 & 49.6 & 38.3 & 28.6 & 44.2 & 7.8 & 38.5 & 52.2 & 23.1 & \textbf{9.2} & 28.1 & 46.8 & 17.1 & 55.5 & 59.6 & 52.7 & 37.9 & 45.0 \\

  13B & $q=20\%$ & \textbf{66.0} & \textbf{54.8} & 42.2 & 31.6 & \textbf{45.8} & \textbf{12.0} & \textbf{42.1} & 54.1 & \textbf{24.9} & 9.0 & \textbf{29.3} & 48.5 &
  17.7 & 58.5 & \textbf{63.8} & 51.2 & \textbf{39.2} & 46.5 \\

  13B & $q=30\%$ & \textbf{66.0} & 53.1 & \textbf{42.9} & \textbf{32.5} & 44.4 & 11.0 & 41.6 & \textbf{54.2} & 24.7 & 8.6 & 29.2 & \textbf{49.0} & \textbf{18.4} &
  \textbf{60.9} & 63.6 & \textbf{52.9} & 38.7 & \textbf{47.2} \\

\bottomrule
\end{tabular}}
\label{tab:quantile_ablation_results}
\end{table*}

\subsection{Partition normalization.}
We additionally evaluate uniform token normalization across the batch by removing the relative per-token upweighting induced by separate partition normalization. For the 7B teacher case, this reduces average reasoning by 5.4 points while improving factual recall modestly by 1 point. This shift is consistent with the reasoning--recall tradeoff observed under increasing distillation strength (\cref{sec:observation}). Notably, reasoning still remains substantially above NTP under uniform normalization, while random routing under our default normalization also substantially underperforms entropy-based routing (\cref{tab:ablation_results}), indicating that both entropy-based routing and separate partition normalization contribute to the reasoning gains.

\subsection{Distributed quantile estimation.} Our default implementation computes the routing quantile over the local micro-batch, and is communication-free by avoiding an additional synchronization point in the training loop. While this overhead is modest relative to model computation at small scale, it becomes less desirable as teacher and student models grow and available memory and communication headroom decrease. In practice, our local micro-batches are sufficiently large and representative of the data mixture such that we observe negligible differences in downstream performance compared with global quantile routing. 

However, this assumption may break down when micro-batch size is constrained to be very small, as may occur with larger teacher and student models. In this setting, one potential alternative is to employ a lagged global quantile. At each optimizer step, we can aggregate teacher entropies over the full global batch and use its $q$-quantile to route the subsequent batch. This decouples quantile estimation from micro-batch size while avoiding a blocking global-quantile computation before routing the current batch.

\section{Experimental Details}
\label{app:exp_details}
\subsection{Training setup}
All our experiments make use of open-source code, checkpoints, and data. We use the \texttt{lingua}~\citep{meta_lingua} framework for pre-training and mid-training, and the \texttt{open-instruct}~\citep{allenai_open_instruct} repository for post-training.

We run pre-training, mid-training, and all stages of post-training except DPO on 32 NVIDIA H200 Tensor Core GPUs across 4 nodes; DPO is conducted on a single node. For efficient distillation, our 7B and 13B teachers are loaded in FP8 quantization, while 1B teachers are kept in BF16; we find that teacher quantization does not substantially affect the predictive entropy ranking of tokens. We set our distillation temperature to $\tau=2$.

\paragraph{Student model architecture.} We provide architecture details for our 1B student model in \cref{tab:student_model_config}.

\begin{table}[t]
\centering
\small
\caption{\textbf{1B student architecture.} We follow the \olmotwo{} 1B model configuration from \citet{walsh2025}.}
\label{tab:student_model_config}
\begin{tabular}{@{}ll@{}}
\toprule
\textsc{Setting} & \textsc{Value} \\
\midrule
Architecture & \olmotwo{} 1B (\texttt{Olmo2ForCausalLM}) \\
Hidden dimension & 2048 \\
Transformer layers & 16 \\
Attention heads & 16 \\
KV heads & 16 (full multi-head attention) \\
FFN intermediate dimension & 8192 \\
Normalization & RMSNorm ($\epsilon=10^{-6}$) \\
QK normalization & Enabled \\
Positional encoding & RoPE ($\theta=500{,}000$) \\
Maximum sequence length & 4096 \\
Weight tying & Disabled \\
Parameters & 1,484,613,632 \\
Vocab Size & 100352 \\
\bottomrule
\end{tabular}
\end{table}

\paragraph{Training hyperparameters.} We provide pre-training and mid-training hyperparameters in \cref{tab:training_hyperparams}. For post-training, we defer hyperparameter choices to the official setup from \citet{walsh2025}, with the exception of a lower learning rate ($5e^{-6}$) during supervised fine-tuning. We find that the catastrophic forgetting of factual knowledge is most pronounced during SFT. Consistent with recommendations from prior work, a smaller learning rate mitigates this loss~\citep{springer2025overtrained}, although it comes at the cost of slightly weaker reasoning performance overall.
\begin{table}[t]
\centering
\small
\caption{\textbf{Hyperparameters for pre-training and mid-training.}}
\label{tab:training_hyperparams}
\begin{tabular}{@{}lcc@{}}
\toprule
\textsc{Setting} & \textsc{Pre-training} & \textsc{Mid-training} \\
\midrule
Training steps & 48{,}000 & 28{,}800 \\
Training tokens & 100.7B & 60.4B \\
Peak learning rate & $4\times10^{-4}$ & $7.45\times10^{-5}$ \\
Learning rate schedule & Cosine & Linear decay \\
Minimum LR ratio & 0.1 & 0.0 \\
Warmup & 4{,}000 steps & None \\
Initialization & Random & \href{https://huggingface.co/allenai/OLMo-2-0425-1B/tree/stage1-step1907359-tokens4001B}{OLMo-2-0425-1B-stage1-4001B} \\
\midrule 
 & \multicolumn{2}{c}{Shared} \\
\midrule
Global batch size &
\multicolumn{2}{c}{2,097,152 tokens/step} \\
Optimizer &
\multicolumn{2}{c}{AdamW} \\
$\beta_1,\beta_2$ &
\multicolumn{2}{c}{(0.9, 0.95)} \\
Weight decay &
\multicolumn{2}{c}{0.1} \\
Gradient clipping &
\multicolumn{2}{c}{1.0} \\
\bottomrule
\end{tabular}
\end{table}

\subsection{Evaluation}

\paragraph{Evaluation tasks.}
\cref{tab:evaluation_tasks} shows the evaluation tasks used in this paper. We adopt the same defaults (e.g., few-shot exemplars, sampling strategy) as \citet{gu-etal-2025-olmes}.

\paragraph{Baselines.}
We set the temperature $T=2$ for all our distillation runs.  In our implementation of \textsc{TRKD}~\citep{goyal2026distilled}, we disable the distillation loss on the 15\% lowest-teacher-entropy tokens, retaining only ground-truth supervision (CE loss) on these tokens (following the original paper's suggestions); the remaining 85\% receive a convex blend $(1-\lambda),\mathrm{CE} + \lambda,T^{2},\mathrm{KL}(p_T|p_S)$. We set the forward KL mixing coefficient $\lambda$ to 0.5. 

\begin{table}[t]
\centering
\small
\caption{Evaluation tasks used for mid-training (\textbf{Mid}) and post-training (\textbf{Post}). We follow the standard evaluation settings used in the \textsc{OLMES} evaluation harness~\citep{gu-etal-2025-olmes}.}
\label{tab:evaluation_tasks}
\setlength{\tabcolsep}{10pt}
\resizebox{\columnwidth}{!}{%
\begin{tabular}{@{}llp{7cm}cc@{}}
\toprule
\textbf{Category} & \textbf{Format} & \textbf{Tasks} & \textbf{Mid} & \textbf{Post} \\
\midrule
\textsc{Reasoning} &
Generation &
\textbf{GSM8K}~\tinycite{cobbe2021trainingverifierssolvemath}, \textbf{GSM-Symbolic}~\tinycite{mirzadeh2025gsmsymbolicunderstandinglimitationsmathematical}, \textbf{GSM-Plus}~\tinycite{li-etal-2024-gsm}, \textbf{BBH}~\tinycite{suzgun2022challenging}, \textbf{DROP}~\tinycite{dua2019dropreadingcomprehensionbenchmark}, \textbf{MATH}~\tinycite{lightman2023lets} &
\checkmark & \checkmark \\
\textsc{Factual recall} &
Generation &
\textbf{TriviaQA}~\tinycite{joshi-etal-2017-triviaqa}, \textbf{Natural Questions}~\tinycite{kwiatkowski-etal-2019-natural}, \textbf{SimpleQA}~\tinycite{wei2024measuringshortformfactualitylarge} &
\checkmark & \checkmark \\
\textsc{Knowledge \& Commonsense} &
Multiple choice &
\textbf{MMLU}~\tinycite{hendrycks2021measuringmassivemultitasklanguage}, \textbf{MMLU-Pro}~\tinycite{mmlu_pro}, \textbf{ARC-Challenge}~\tinycite{clark2018thinksolvedquestionanswering}, \textbf{OpenBookQA}~\tinycite{mihaylov2018suitarmorconductelectricity}, \textbf{WinoGrande}~\tinycite{sakaguchi2019winograndeadversarialwinogradschema}, \textbf{AGIEval}~\tinycite{zhong2023agievalhumancentricbenchmarkevaluating} & \checkmark
& \checkmark \\
\textsc{Instruction Following} &
Generation &
\textbf{IFEval}~\tinycite{zhou2023instructionfollowingevaluationlargelanguage} &
& \checkmark \\
\bottomrule
\end{tabular}
}
\end{table}

\subsection{Analysis methodology}

\paragraph{Teacher supervision analysis.} 
For each teacher we score the same 240 Dolmino documents (40 per domain, sampled with a fixed seed), yielding 107,555 token-level next-token predictions per teacher, decomposed by domain as 25,981 from DCLM, 22,514 from Wikipedia, 16,692 from StackExchange, 16,599 from PeS2o, 12,886 from Math, and 12,883 from FLAN. All three teachers therefore probe identical positions; in other words, between- and within-teacher comparisons share the same support.

\paragraph{Gold-answer gradient analysis.}
\label{app:gold_answer_gradient_derivation}
In \cref{subsec:gradient_analysis}, we analyze the gradients induced by the NTP, FKD, and RKD training objectives with respect to the ground-truth token. For FKD and RKD, we evaluate $\alpha=0.5$; thus, the dashed $0.5\times$ NTP level in \cref{fig:gold_attenuation} corresponds to the $1-\alpha$ floor as the distillation-gradient contribution vanishes.

We provide full derivations for each objective. Let $A$ denote the set of accepted first-token IDs obtained from the gold answer aliases (e.g., if a prompt is "Name a European capital: ", then $A$ might include the token IDs corresponding to \{"Paris", "London", "Berlin"\}). We define the \emph{gold-answer gradient} $g$ as the negative gradient of some loss $\mathcal{L}$ with respect to the student logits corresponding to the set $A$:
\begin{align}
g &= -\frac{\partial \mathcal{L}}{\partial z_{S, A}},
\quad \text{where}\quad
\frac{\partial}{\partial z_{S, A}} =
\sum_{y\in A}
\frac{\partial}{\partial z_{S, y}}.
\end{align}
Intuitively, $g$ is a scalar that measures the strength of the learning signal induced by $\mathcal{L}$ on all accepted answer tokens in $A$; larger values of $g$ correspond to stronger pressure to increase the probability of the correct answer. We next derive the closed forms of $g$ for the training objectives in our study: $g_{\mathrm{NTP}}$, $g_{\mathrm{FKD}}$, and $g_{\mathrm{RKD}}$.

\paragraph{Deriving $g_{\mathrm{NTP}}$.}
Recall that the NTP objective is the standard cross-entropy loss ($\mathcal{L}_{\mathrm{CE}}$). 
Let $p_S^\tau = \operatorname{softmax}(z_S/\tau)$ denote the student distribution at temperature $\tau$. The cross-entropy term is computed at $\tau{=}1$ against the observed next token $y^\star$. By the construction of our probe, we guarantee that $y^\star \in A$. Since $\mathcal{L}_{\mathrm{CE}} = -\log p_S^1(y^\star)$ and using the standard Jacobian of the log-softmax, $\frac{\partial \log p_S^1(y^\star)}{\partial z_{S,j}} =\delta_{jy^\star}-p_S^1(j)$ for any arbitrary student logit $z_{S,j}$,
\begin{align}
g_{\mathrm{NTP}}
&=
-\sum_{y\in A}
\frac{\partial \mathcal L_{\mathrm{CE}}}{\partial z_{S,y}}\\
&=
-\sum_{y\in A}
\left(p_S^1(y)-\delta_{yy^\star}\right) \\
&=
-\sum_{y\in A}p_S^1(y)
+
\sum_{y\in A}\delta_{yy^\star} \\
&=
1-p_S^1(A).
\end{align}
Intuitively, this gradient represents the probability mass still missing from the accepted answer set. The cross-entropy loss pulls these logits upward until $p_S^1(A) = 1$. Note that this target is absolute and does not depend on any reference teacher.

\paragraph{Deriving $g_{\mathrm{FKD}}$.} We next examine the FKD objective, which consists of a convex combination of the forward KL (FKL) loss and the cross-entropy loss as weighted by $\alpha \in [0,1]$. 

Let us tackle the FKL loss first, which depends on a reference teacher $T$. Recall that the FKL equation is, by definition, $\mathrm{KL}(p_T^\tau \| p_S^\tau) = \sum_y p_T^\tau(y)\log p_T^\tau(y) - \sum_y p_T^\tau(y)\log p_S^\tau(y)$, with $p_T^\tau$ as the teacher distribution and $p_S^\tau$ as the student distribution, both scaled by temperature $\tau$. Since the teacher distribution $p_T^\tau$ is fixed with respect to the student logits $z_S$, the first term is constant and has zero derivative, and only the second term $-\sum_y p_T^\tau(y)\log p_S^\tau(y)$ contributes to the gradient.

Using the standard Jacobian of the log-softmax,
$\frac{\partial \log p_S^\tau(y)}{\partial z_{S,j}} = \frac{1}{\tau}\big(\delta_{yj} - p_S^\tau(j)\big)$, and the observation that
$\sum_y p_T^\tau(y) = 1$,
\begin{align}
\frac{\partial\, \mathrm{KL}(p_T^\tau\|p_S^\tau)}{\partial z_{S,j}}
&= -\sum_y p_T^\tau(y)\,\frac{1}{\tau}\big(\delta_{yj} - p_S^\tau(j)\big) \\
&= \frac{1}{\tau}\big(p_S^\tau(j) - p_T^\tau(j)\big).
\end{align}
Summing the negative gradients over $y \in A$ and combining with the cross-entropy term in the full FKD objective
$\mathcal{L}_{\mathrm{FKD}} = (1{-}\alpha)\mathcal{L}_{\mathrm{CE}} + \alpha\tau^2\,\mathrm{KL}(p_T^\tau\|p_S^\tau)$, where the
$\tau^2$ cancels one factor of $1/\tau$, leads to
\begin{align}
g_{\mathrm{FKD}}
&= -\sum_{y\in A} \frac{\partial \mathcal L_{\mathrm{FKD}}}{\partial z_{S,y}} \\
&= \underbrace{(1-\alpha)\big(1 - p_S^1(A)\big)}_{\text{CE term}} + \underbrace{\alpha\tau\big(p_T^\tau(A) - p_S^\tau(A)\big)}_{\text{FKL term}}. \label{eq:fkd}
\end{align}

Comparing against $g_{\mathrm{NTP}}$, the FKL term replaces the cross-entropy target of $1$ with $p_T^{\tau}(A)$, the teacher's probability mass on the accepted answer set. Thus, unlike cross-entropy, the distillation component does not continue increasing the student's accepted-answer mass once it reaches the teacher's: when $p_S^{\tau}(A) = p_T^{\tau}(A)$, the FKL contribution vanishes. 

When $p_S^{\tau}(A) > p_T^{\tau}(A)$, it becomes negative and actively pushes the student back toward the teacher distribution. In the full FKD objective (\cref{eq:fkd}), this negative distillation gradient competes with the remaining positive cross-entropy gradient and can dominate when the teacher assigns substantially less mass to $A$ than the student. This regime may be especially relevant during mid-training, when the student may already assign high probability to factual continuations for which the teacher remains uncertain. 

\paragraph{Deriving $g_{\mathrm{RKD}}$.}
We now examine the RKD objective, which substitutes the Forward KL divergence with the Reverse KL divergence, $\mathrm{RKL}(p_S^\tau \| p_T^\tau) = \sum_y p_S^\tau(y) \log \frac{p_S^\tau(y)}{p_T^\tau(y)}$.
To find the gradient with respect to any arbitrary student logit $z_{S,j}$, we apply the product rule and the softmax Jacobian $\frac{\partial p_S^\tau(y)}{\partial z_{S,j}} = \frac{1}{\tau}p_S^\tau(y)\big(\delta_{yj} - p_S^\tau(j)\big)$:
\begin{align}
\frac{\partial\,\mathrm{RKL}(p_S^\tau \| p_T^\tau)}{\partial z_{S,j}}
&= \sum_y \frac{\partial p_S^\tau(y)}{\partial z_{S,j}} \left(\log \frac{p_S^\tau(y)}{p_T^\tau(y)} + 1 \right) \\
&= \frac{1}{\tau} \sum_y p_S^\tau(y)\big(\delta_{yj} - p_S^\tau(j)\big) \left(\log \frac{p_S^\tau(y)}{p_T^\tau(y)} + 1 \right).
\end{align}
Because the sum of probabilities is 1, the gradient of that sum is zero ($\sum_y \frac{\partial p_S^\tau(y)}{\partial z_{S,j}} \cdot 1 = 0$), causing the $+1$ term to vanish. Distributing the remaining terms yields:
\begin{align}
\frac{\partial\,\mathrm{RKL}(p_S^\tau \| p_T^\tau)}{\partial z_{S,j}}
&= \frac{1}{\tau} \left[ p_S^\tau(j) \log \frac{p_S^\tau(j)}{p_T^\tau(j)} - p_S^\tau(j) \sum_y p_S^\tau(y) \log \frac{p_S^\tau(y)}{p_T^\tau(y)} \right] \\
&= \frac{1}{\tau} p_S^\tau(j) \Big( \log \frac{p_S^\tau(j)}{p_T^\tau(j)} - \mathrm{RKL}(p_S^\tau \| p_T^\tau) \Big).
\end{align}
Summing the negative gradients over $y \in A$ and combining with the cross-entropy term under the full RKD objective $\mathcal{L}_{\mathrm{RKD}} = (1{-}\alpha)\mathcal{L}_{\mathrm{CE}} + \alpha\tau^2\,\mathrm{RKL}(p_S^\tau\|p_T^\tau)$, we obtain:
\begin{align}
g_{\mathrm{RKD}}
&= -\sum_{y\in A} \frac{\partial \mathcal L_{\mathrm{RKD}}}{\partial z_{S,y}} \\
&= \underbrace{(1-\alpha)\big(1 - p_S^1(A)\big)}_{\text{CE term}} - \underbrace{\alpha\tau \sum_{y \in A} p_S^\tau(y) \Big( \log \frac{p_S^\tau(y)}{p_T^\tau(y)} - \mathrm{RKL}(p_S^\tau \| p_T^\tau) \Big)}_{\text{RKL term}}. \label{eq:rkd}
\end{align}

Note that in \cref{eq:rkd}, RKL can likewise exert active negative pressure on ground-truth answer logits rather than merely reducing their positive supervision. The contribution for an accepted answer token $y$ is negative to $g_{\mathrm{RKD}}$ whenever its log-probability ratio $\log \frac{p_S^\tau(y)}{p_T^\tau(y)}$ exceeds the student-averaged log ratio $\mathrm{RKL}(p_S^\tau \| p_T^\tau)$. 

Thus, although their respective gradient structure may differ, both KL directions can penalize student predictions on the ground truth answers if they exceed the teacher's relative preference. 

\subsection{Standalone student and teacher results}

For reference, we report student (\olmotwo{} 1B, after pre-training but before mid-training) and teacher (\olmotwo{} 1B, 7B, and 13B Instruct) performance in \cref{tab:teacher_performance}. We remark that the 1B student already matches the 1B Instruct teacher on factual recall, despite substantially weaker reasoning performance, suggesting that much of the teacher’s advantage at the start of mid-training lies in procedural capabilities rather than factual knowledge. The larger teachers, however, consistently outperform the student across all evaluated capabilities, indicating that the observed reasoning–recall tradeoff cannot simply be attributed to teachers lacking the factual knowledge being learned.

\begin{table*}[t]
\centering
\tiny
\setlength{\tabcolsep}{2.8pt}
\renewcommand{\arraystretch}{1.15}
\caption{
\textbf{Downstream performance of the student initialization and teacher models.}
We report performance of the \olmotwo{} 1B Stage 1 student initialization and
the \olmotwo{} 1B, 7B, and 13B Instruct teachers on the downstream evaluation suite.
Benchmark names are abbreviated; see \cref{tab:evaluation_tasks} for full task names.
}
\vspace{-10pt}
\setlength{\tabcolsep}{1pt}
\resizebox{\columnwidth}{!}{%
\begin{tabular}{l|cccccc|ccc|cccccc}
\toprule
&
\multicolumn{6}{c|}{Reasoning}
&
\multicolumn{3}{c|}{Factual Recall}
&
\multicolumn{6}{c}{Knowledge \& Commonsense} \\
\cmidrule(lr){2-7}
\cmidrule(lr){8-10}
\cmidrule(l){11-16}
Model
& \textbf{GSM8K} & \textbf{GSM-S} & \textbf{GSM+} & \textbf{BBH} & \textbf{DROP} & \textbf{MATH}
& \textbf{TQA} & \textbf{NQ} & \textbf{SQA}
& \textbf{MMLU} & \textbf{MMLU-P} & \textbf{ARC-C} & \textbf{OBQA} & \textbf{Wino} & \textbf{AGI} \\
\midrule

\olmotwo{} \textsc{1B Stage 1}
& 3.2 & 1.9 & 2.1 & 29.9 & 24.8 & 1.2
& 52.5 & 21.0 & 8.0
& 26.9 & 11.3 & 26.2 & 26.2 & 50.6 & 23.9 \\

\midrule[0.5pt]

\olmotwo{} \textsc{1B-Inst}
& 67.3 & 49.1 & 42.7 & 35.2 & 32.3 & 9.0
& 50.8 & 21.1 & 7.5
& 41.9 & 17.8 & 47.2 & 48.8 & 50.4 & 35.0 \\

\olmotwo{} \textsc{7B-Inst}
& 79.2 & 69.6 & 57.0 & 50.4 & 59.2 & 17.4
& 75.8 & 38.1 & 10.6
& 61.8 & 32.4 & 80.1 & 78.4 & 66.9 & 50.1 \\

\olmotwo{} \textsc{13B-Inst}
& 82.9 & 72.6 & 59.9 & 58.2 & 70.4 & 22.0
& 83.2 & 49.4 & 10.9
& 66.7 & 37.2 & 83.4 & 85.0 & 71.3 & 54.0 \\

\bottomrule
\end{tabular}}
\label{tab:teacher_performance}
\end{table*}

\section{Supplementary Analyses}
\label{app:more_analysis}
\subsection{How generalizable is the mid-training tradeoff across model families?}
\label{app:smollm}
Our main experiments use the \olmotwo{} model family and training pipeline. To test whether this observed mid-training tradeoff is generalizable, we move off \olmotwo{} entirely and repeat our tradeoff analysis with SmolLM: namely, a substantially smaller SmolLM2 360M student, a SmolLM2 1.7B Instruct teacher~\citep{allal2025smollm}, and SmolLM3 Stage-3 training data~\citep{smollm3}. In line with our main experiments, we train for 100B tokens during pre-training and 60B tokens during mid-training, using the same SmolLM3 Stage-3 data in both stages, and retain the same evaluation setup. We also include \ours{} at mid-training (keeping the same routing threshold $q=20\%$). 

\begin{figure*}[t]
    \centering
    \includegraphics[width=\textwidth]{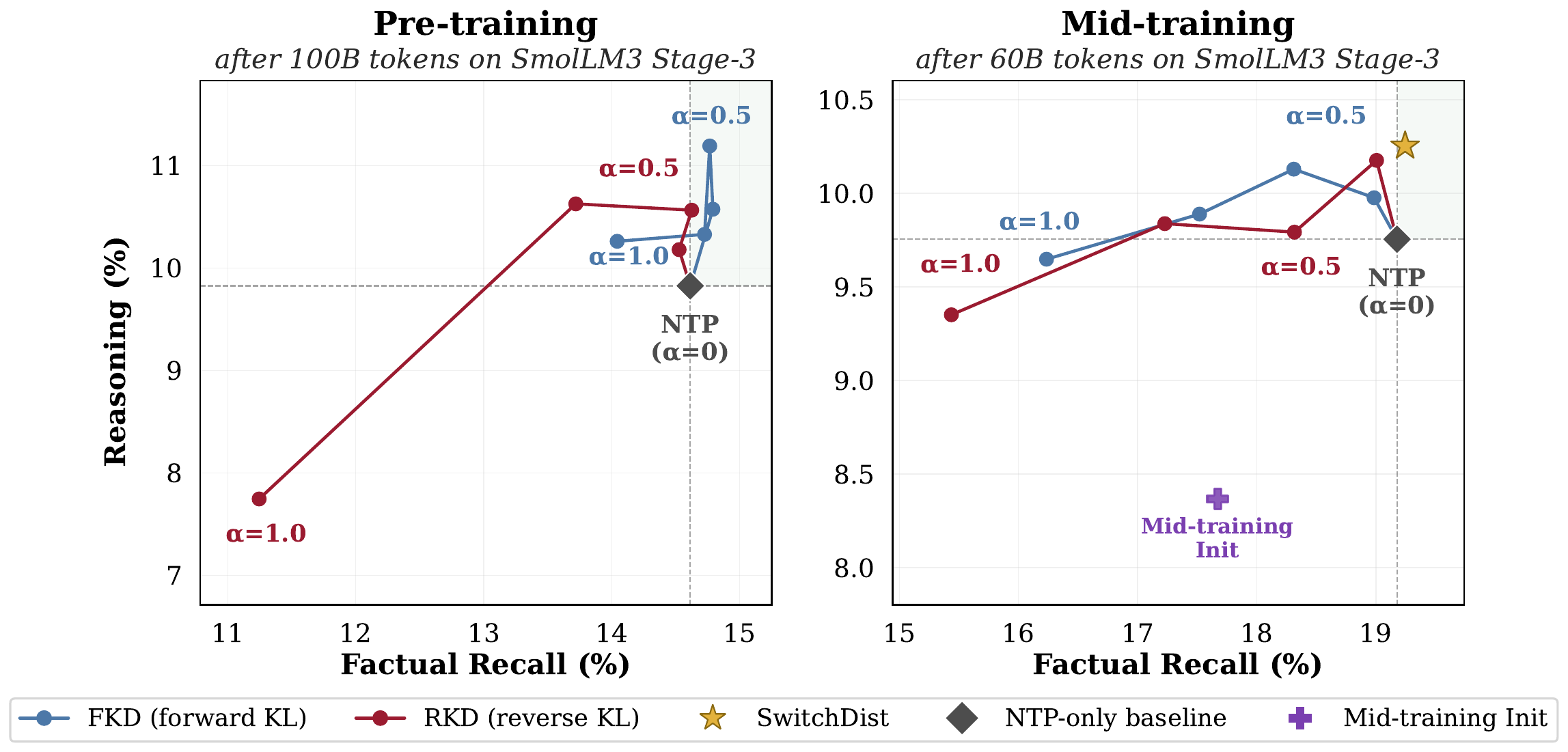}
    \vspace{-16pt}
    \caption{
  \textbf{Reasoning-recall tradeoff using the SmolLM ecosystem.}
    }
    \label{fig:tradeoff_smollm2}
\end{figure*}

Despite noisier and less monotonic tradeoff curves at this substantially smaller scale, we recover the same qualitative behavior (\cref{fig:tradeoff_smollm2}): during pre-training, moderate forward KL distillation admits Pareto improvements in both reasoning and factual recall over NTP. During mid-training, however, standard forward and reverse KL improve reasoning only at the expense of factual recall. \ours{} is able to mitigate this tradeoff \emph{at mid-training}, jointly improving reasoning and factual recall over NTP. 

These results suggest that the stage-dependent behavior of distillation, as well as the benefit of \ours{}, extends beyond the \olmotwo{} model and data ecosystem. 

\subsection{How generalizable is teacher supervision asymmetry?}
\label{app:teacher_supervision_asymmetry}

One of our main findings is that teacher supervision is asymmetric across domains using the \olmotwo{} Instruct models as teachers; specifically, tokens from procedural domains (math, instruction-following) tend to have lower teacher entropy than those from knowledge-intensive domains. Here, we ask whether this phenomenon generalizes across training stages and model families.  

\begin{figure*}[!t]
    \centering
    \includegraphics[width=\textwidth]{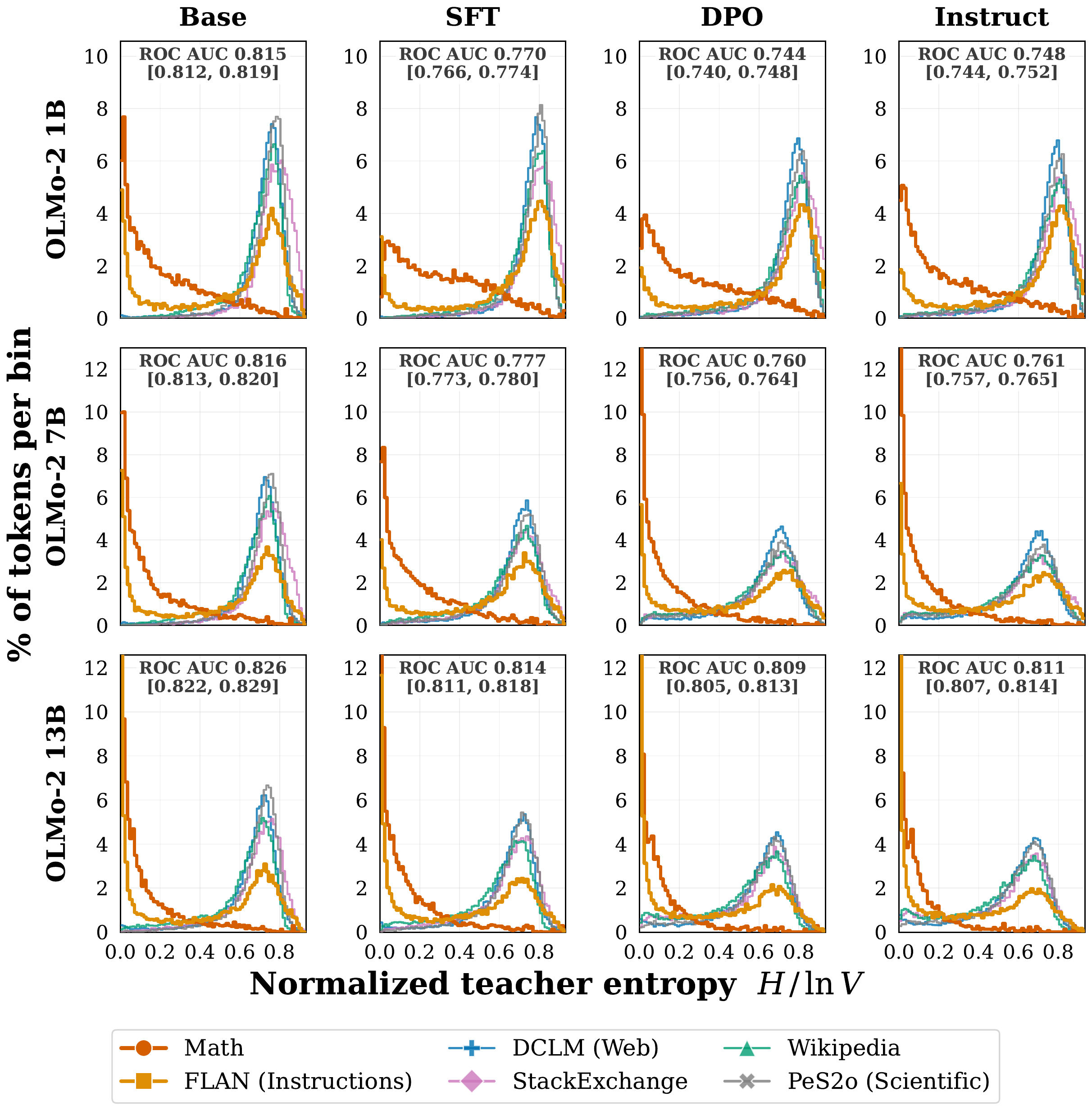}
  \vspace{-16pt}
    \caption{
  \textbf{Asymmetric teacher supervision holds across training stages.}
    }\label{fig:teacher_supervision_training_stage}
\end{figure*}

\paragraph{Across training stages.}
\label{app:training_stages}
\Cref{fig:teacher_supervision_training_stage} repeats the analysis using each teacher model at different stages of training and parameter size. To quantify the separation, we report the receiver operating characteristic area under the curve (ROC AUC) with 95\% CI. The ROC AUC quantifies how well teacher entropy distinguishes procedural (Math and FLAN) from knowledge-intensive (DCLM, Wikipedia, StackExchange, and PeS2o) domains using the teacher predictive entropy of tokens. More specifically, an ROC AUC of $x$ indicates that a randomly selected procedural token has lower teacher entropy than a randomly selected knowledge-intensive token with probability $x$; $0.5$ is random chance.

Across all 12 stage-size cells, the ROC AUC stays within
$[0.744,\,0.826]$ (with 95\% CI width ${\le}\,0.005$), i.e., the entropy gap between procedural and knowledge-intensive tokens is a pretraining-era property that neither SFT, DPO, nor the
final Instruct stage removes; the small monotone erosion visible from Base to
Instruct at every size (e.g.\ $0.816\!\to\!0.761$ at 7B) is the only stage-level
effect and never approaches chance.

\paragraph{Across model families.}
\label{app:model_families}
To test whether this finding generalizes across other model families, we additionally repeat our analysis on OLMo-3 7B Instruct~\citep{olmo2026olmo3}, Qwen 3 8B~\citep{yang2025qwen3technicalreport}, Gemma 3 12B Instruct~\citep{gemmateam2025gemma3technicalreport}, and Granite 3.3 8B Instruct~\citep{granite33}, all of which are recent open-weight instruction-tuned models. As these models have different vocabularies, we report the entropy normalized by the vocabulary size. 

\begin{figure*}[!t]
    \centering
    \includegraphics[width=\textwidth]{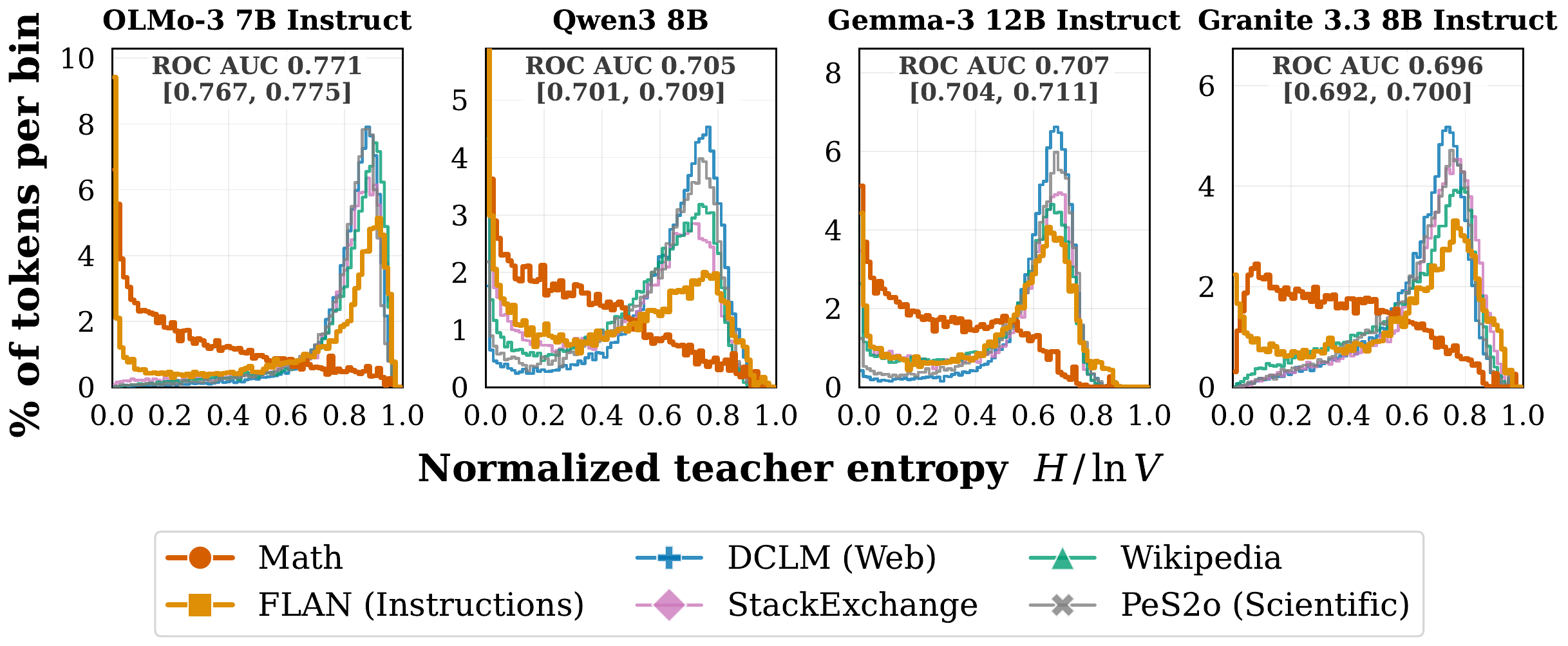}
    \vspace{-16pt}
    \caption{
  \textbf{Asymmetric teacher supervision holds across model families.}
    }
    \label{fig:teacher_supervision_crossfamily}
\end{figure*}

In \cref{fig:teacher_supervision_crossfamily}, all four exhibit the same qualitative shape as \olmotwo{}: procedural mass concentrates at the low end of normalized entropy, and knowledge-intensive mass at the high end, and every AUC sits well above chance ($0.696$ to $0.771$; 95\% CI widths ${\le}\,0.008$). The separation is thus not an \olmotwo{}-specific artifact but a general property of modern instruction-tuned models.

\subsection{Robustness across teacher sizes}
\label{app:analyses_teacher_sizes}

\begin{figure*}[t]
    \centering
    
    \includegraphics[width=\textwidth]{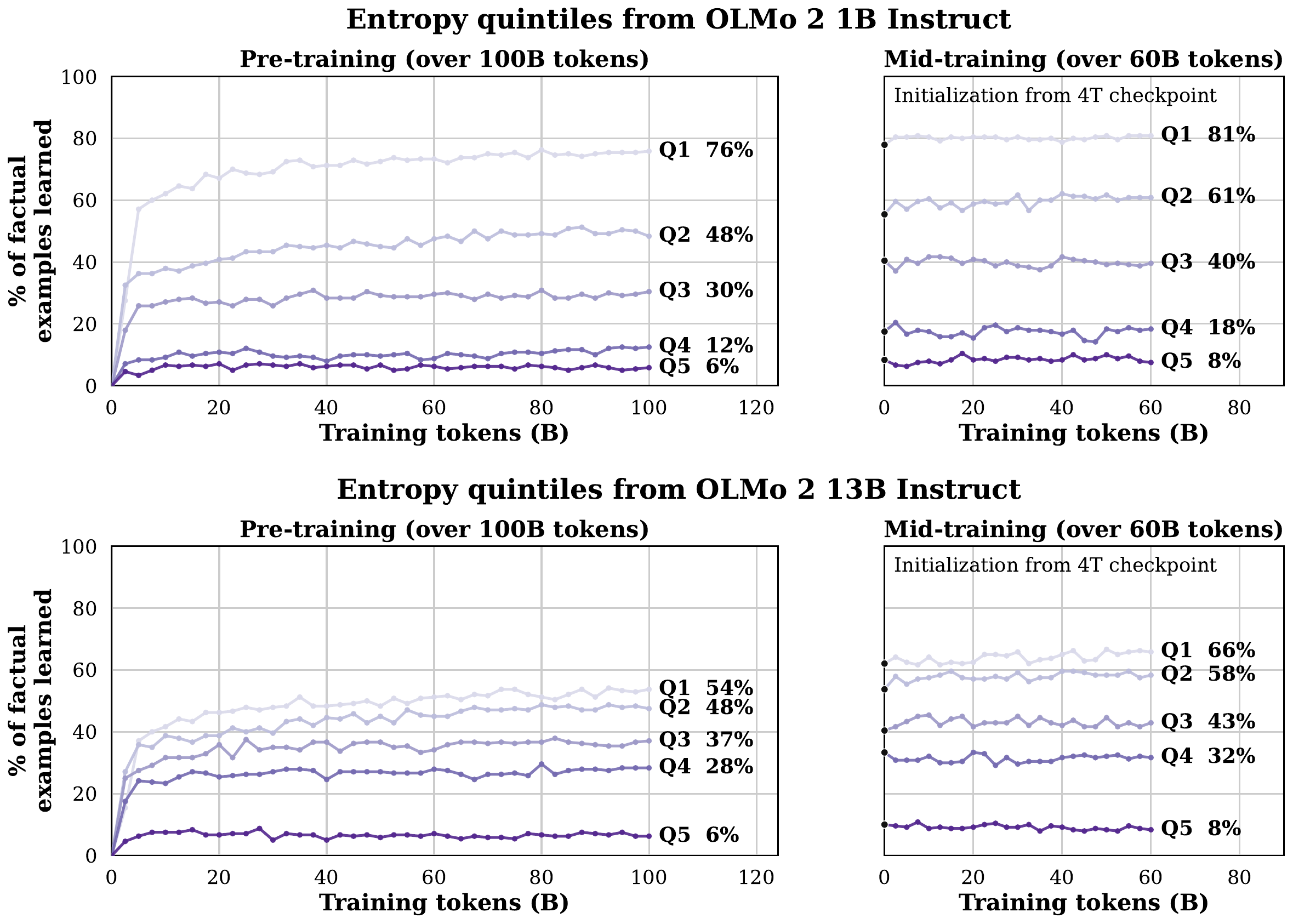}
    \vspace{-16pt}
    \caption{
  \textbf{Teacher entropy predicts factual acquisition under standard NTP.} The same qualitative trend holds when using \olmotwo{} 1B and 13B Instruct as teachers.
    }
     \label{fig:quintile_acquisition_1b13b}
\end{figure*}

\paragraph{Factual acquisition over the course of training.} In \cref{subsec:student_side}, we showed that a correlation between predictive entropy and factual acquisition exists using  \olmotwo{} 7B Instruct as a teacher. In \cref{fig:quintile_acquisition_1b13b}, we find that the same empirical trend persists across teacher sizes, using \olmotwo{} Instruct 1B and 13B as teachers.

\begin{figure*}[t]
    \centering
    \includegraphics[width=\textwidth]{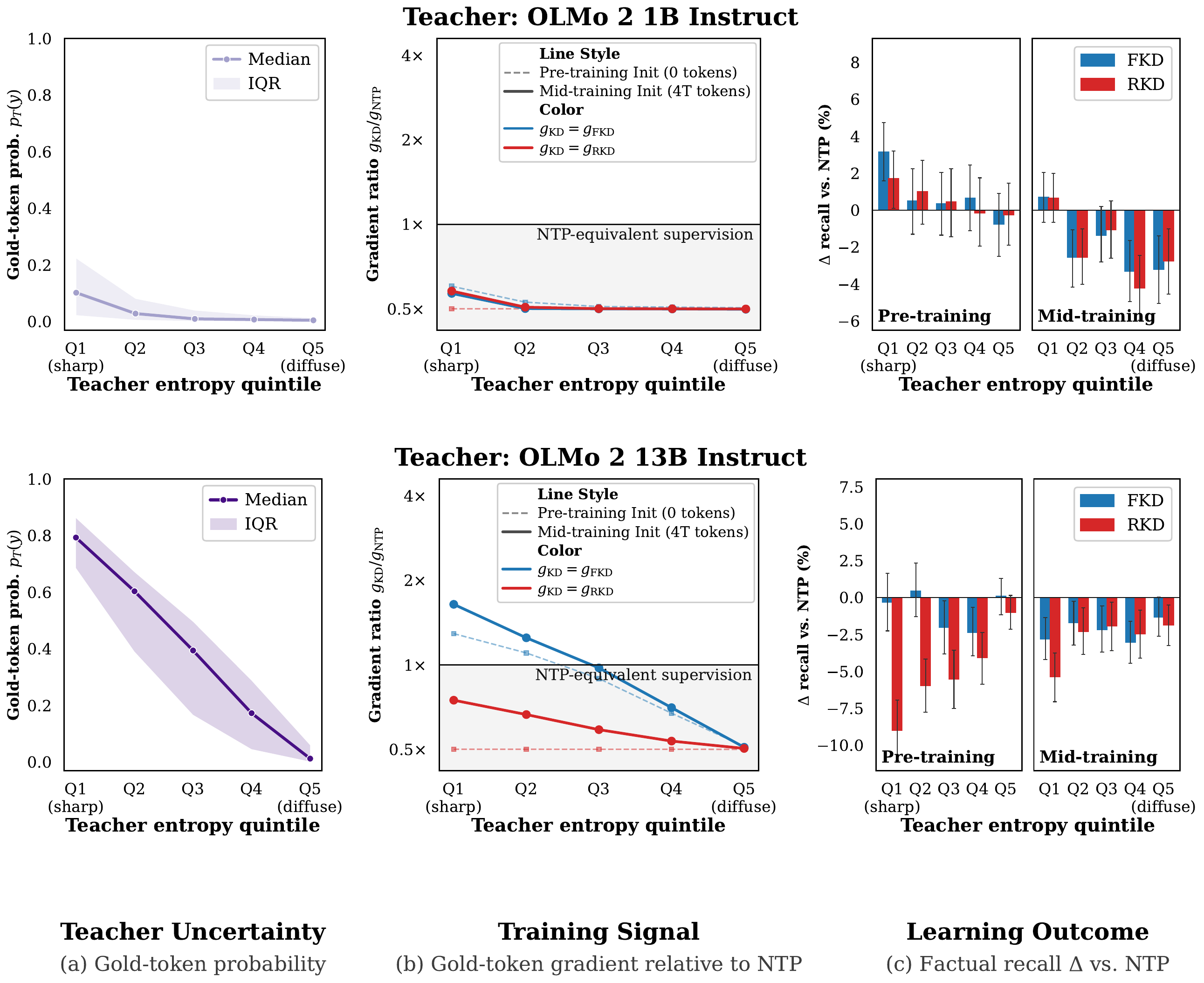}
    \vspace{-16pt}
    \caption{
  \textbf{KD analysis on factual recall examples.} The same qualitative trends hold when using \olmotwo{} 1B and 13B Instruct as teachers, in accord with using the 7B teacher in \cref{fig:gold_attenuation}.
    }
    \label{fig:gold_attenuation_1b13b}
\end{figure*}

\paragraph{Factual recall analysis of KD.}
In a similar vein, we show that the KD analysis on factual recall examples is also largely consistent for \olmotwo{} 1B and 13B Instruct teachers in \cref{fig:gold_attenuation_1b13b}. 

\subsection{Forward and reverse KL exhibit different optimization geometry.}
Forward and reverse KD differ only in the choice of divergence, yet consistently produce different reasoning--recall tradeoff patterns. To better understand this difference, we analyze how the KL divergence loss in the objective interacts with the CE component. Specifically, at fixed model parameters, we compute the cosine similarity between the closed-form logit gradients of the CE and KL losses (e.g., FKL and RKL) on held-out pretraining data. Higher cosine values indicate stronger alignment between the two objectives.

\begin{figure*}[t]
    \centering
    \includegraphics[width=\textwidth]{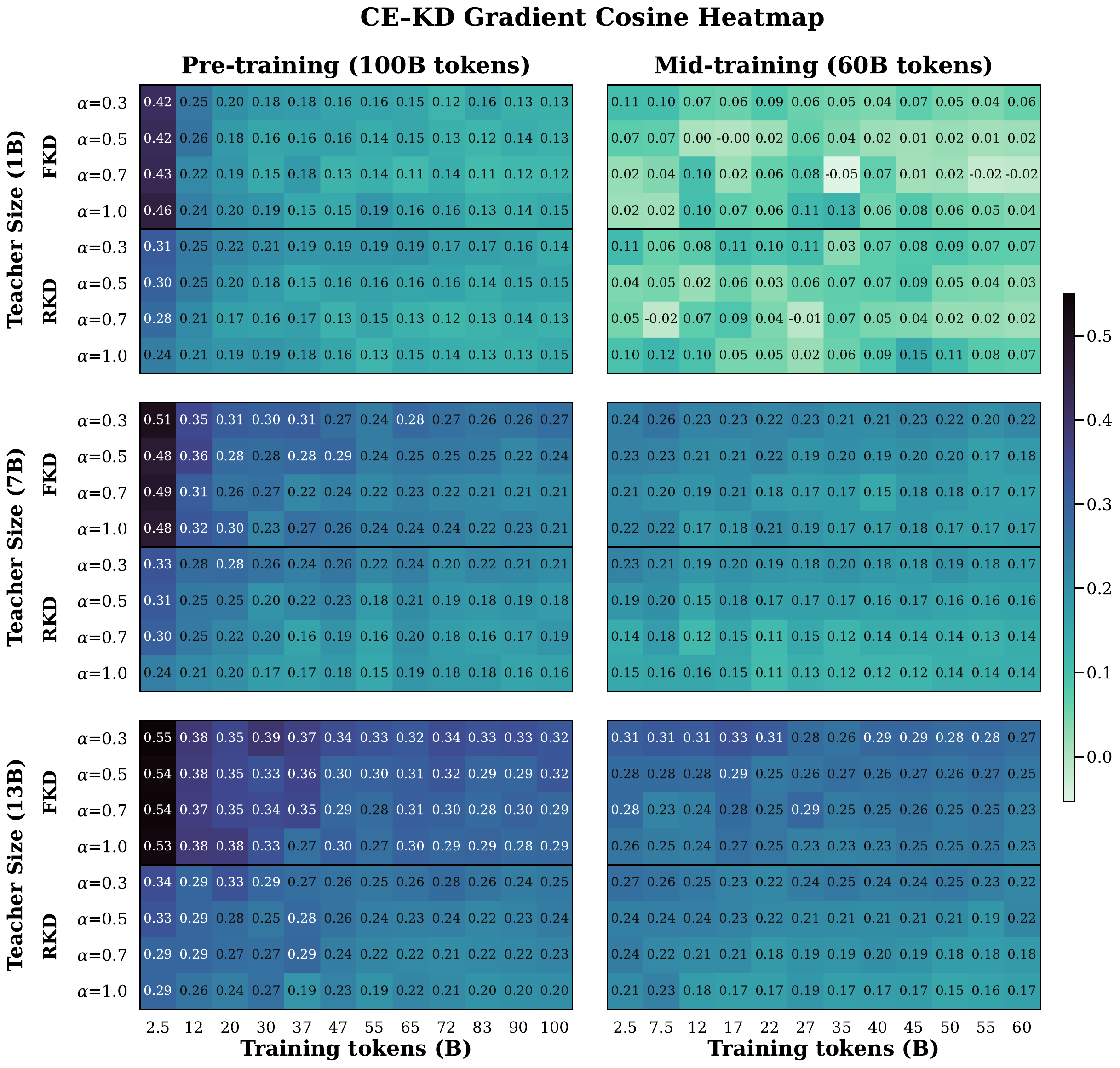}
    \vspace{-16pt}
    \caption{
  \textbf{FKL exhibits higher gradient alignment with the CE objective than RKL throughout training.}
    }
    \label{fig:ce_kl_cosine_heatmap}
\end{figure*}

\cref{fig:ce_kl_cosine_heatmap} reports the resulting gradient cosine throughout training across KD mixture weights. At larger teacher sizes (7B and 13B) especially, FKL consistently exhibits higher CE–KL gradient alignment than RKL during both pre-training and mid-training, with the difference persisting across training checkpoints and mixture weights. This indicates that, locally, the FKL objective favors update directions more similar to NTP than does RKL. Consequently, FKL perturbs the next-token prediction optimization direction less than RKL, providing one explanation for why the two KD objectives occupy systematically different positions on the reasoning–factual recall frontier. Notably, this distinction depends on teacher size: with the size-matched 1B teacher, FKL shows greater alignment than RKL primarily during early pre-training, while the difference largely disappears later in pre-training and throughout mid-training. 

\subsection{\ours{} accelerates mid-training reasoning acquisition}

\begin{figure*}[t]
    \centering
    \includegraphics[width=\textwidth]{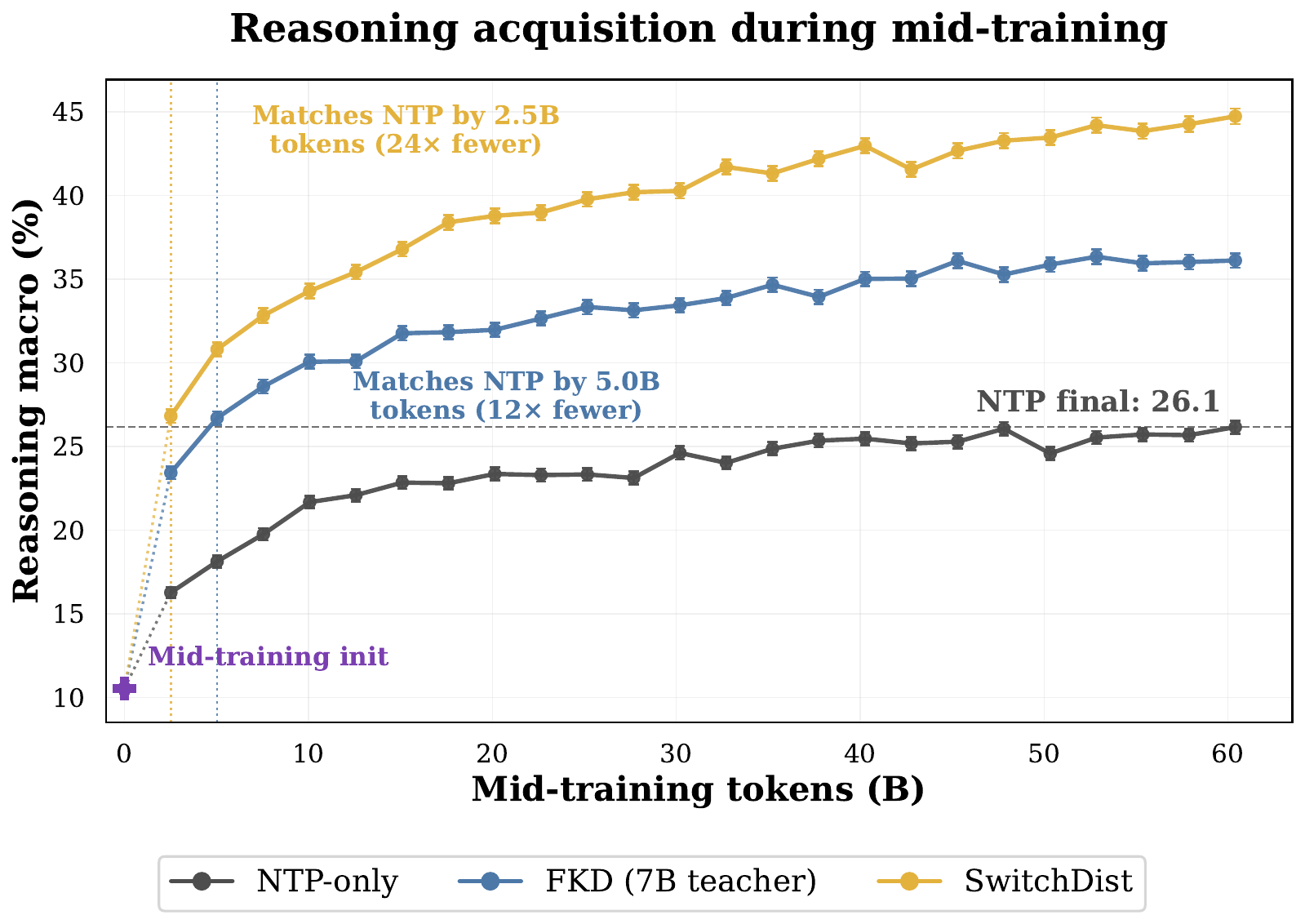}
\vspace{-16pt}
\caption{\textbf{\ours{} substantially accelerates reasoning acquisition during mid-training.}
\ours{} surpasses the final reasoning performance of the 60B-token NTP baseline within the first 2.5B mid-training tokens evaluated, while standard forward KD does so with double the amount (5B tokens). Both distillation methods continue to improve with additional training, maintaining a substantial advantage over NTP throughout mid-training.
    }
    \label{fig:plot_trajectory}
\end{figure*}

Our main experiments compare methods at the end of mid-training. Here, we instead examine their learning trajectories to understand \emph{when} reasoning gains emerge. We evaluate intermediate checkpoints throughout the 60B-token mid-training run, comparing standard NTP, forward KD (at $\alpha=0.5$), and \ours{}, using the \olmotwo{} 7B Instruct teacher. \cref{fig:plot_trajectory} reports macro-averaged reasoning performance as a function of the number of mid-training tokens consumed.

We observe that the reasoning advantage from distillation emerges remarkably early: the NTP baseline reaches a final reasoning macro-average of 26.1\% after 60B mid-training tokens. In contrast, \ours{} exceeds this level by the first evaluated checkpoint at 2.5B tokens, corresponding to $1/24$ of the NTP training token budget. FKD reaches the same threshold with twice that budget ($1/12$). Notably, these early gains do not simply reflect faster convergence to the same solution: both FKD and \ours{} continue to improve throughout training and finish substantially above the NTP baseline, with \ours{} maintaining the strongest reasoning performance across the trajectory.  

\subsection{\ours{} improves pass@$k$ at low sampling budgets}

\begin{figure*}[t]
    \centering
    \includegraphics[width=\textwidth]{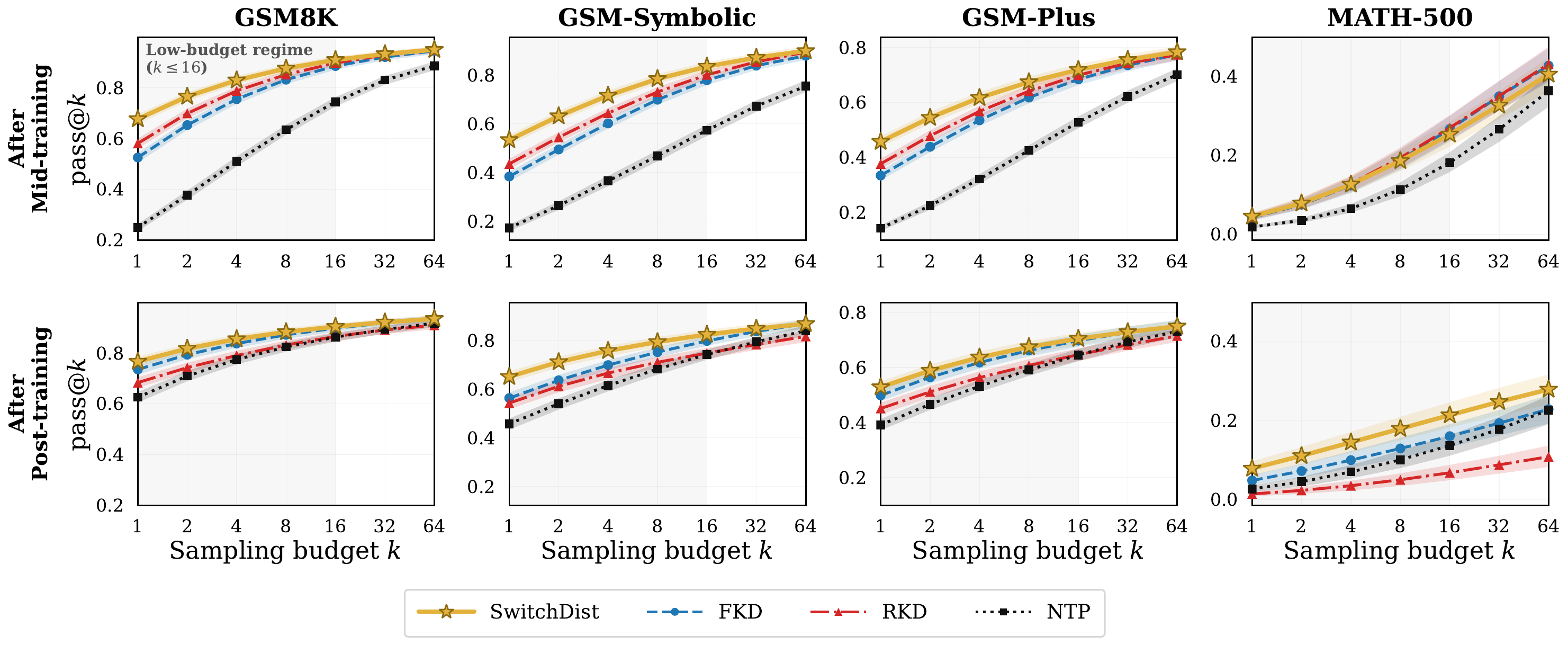}
\vspace{-16pt}
\caption{\textbf{\ours{} improves pass@$k$ performance for reasoning tasks (GSM8K, GSM-Symbolic, GSM-Plus, Math-500), particularly at low sampling budgets.} 
We report pass@$k$ performance for $k \in \{1,2,4,8,16,32,64\}$ after mid-training (top) and post-training (bottom).}
    \label{fig:passk}
\end{figure*}

\cref{fig:passk} examines whether the reasoning gains from \ours{} persist beyond pass@1 as additional inference-time samples become available. After mid-training, \ours{} and the other KD methods consistently outperform NTP across sampling budgets. On the GSM tasks, \ours{} performs best among all methods, with its largest advantages in the low-budget regime ($k \leq 16$). On MATH-500, \ours{} slightly trails FKD and RKD. The differences among KD methods progressively shrink as $k$ increases, suggesting that additional test-time compute can partially compensate for weaker per-sample reasoning performance.

After post-training, the KD methods become substantially closer on GSM8K, GSM-Symbolic, and GSM-Plus, consistent with post-training narrowing the reasoning gaps observed after mid-training. Nevertheless, \ours{} maintains a slight lead across the GSM tasks. Notably, the trend reverses on MATH-500: \ours{} overtakes FKD and achieves the strongest pass@$k$ performance across all sampling budgets, with its advantage persisting through $k=64$. Overall, these results suggest that \ours{} is particularly effective under constrained inference-time sampling budgets, while its gains can persist even at larger budgets on more challenging reasoning tasks.

\subsection{\ours{} improves code generation performance}
\begin{table}[t]
\centering
\tiny
\setlength{\tabcolsep}{4pt}
\renewcommand{\arraystretch}{1.15}
\caption{
\textbf{Code generation results on MBPP~\citep{austin2021programsynthesislargelanguage} after mid-training.}
Following the OLMES evaluation setup, we report pass@1 (\%) on MBPP (500 samples).
$\pm$ denotes the 95\% confidence interval, and stars indicate statistically significant differences from the shared NTP
baseline under paired per-problem tests
($^{*}p<0.05$, $^{**}p<0.01$, $^{***}p<0.001$).
}
\vspace{-10pt}
\resizebox{0.4\columnwidth}{!}{%
\begin{tabular}{ll|c}
\toprule
$T$ & Method & MBPP ($n=500$) \\
\midrule
N/A & NTP & 4.80 $\pm$ 1.87 \\
7B & FKD & 7.60 $\pm$ 2.32$^{*}$ \\
7B & RKD & 7.80 $\pm$ 2.35$^{*}$ \\
7B & TRKD & 5.60 $\pm$ 2.02 \\
7B & \textbf{SD} & \textbf{8.00 $\pm$ 2.38}$^{*}$ \\
\midrule
N/A & NTP & 4.80 $\pm$ 1.87 \\
13B & FKD & 8.60 $\pm$ 2.46$^{**}$ \\
13B & RKD & 9.60 $\pm$ 2.58$^{***}$ \\
13B & TRKD & 6.80 $\pm$ 2.21 \\
13B & \textbf{SD} & \textbf{10.00 $\pm$ 2.63}$^{***}$ \\
\bottomrule
\end{tabular}%
}
\label{tab:code_generation_results}
\end{table}

We additionally provide code generation results for our mid-training settings on Mostly Basic Python Problems~\citep{austin2021programsynthesislargelanguage}. We do not include code generation in our main evaluation because explicitly code-related data constitutes a relatively small portion of the \olmotwo{} mid-training mixture; for example, StackExchange's sampling ratio is only ${\sim}2.5\%$ at mid-training. Nevertheless, for comprehensiveness, we evaluate on MBPP to test whether the observed mid-training gains in procedural reasoning extend to code generation.

As \cref{tab:code_generation_results} shows, all distillation settings improve over NTP, with \ours{} achieving the highest pass@1 for both teacher sizes. The improvements from \ours{} over NTP are statistically significant, while the smaller differences among distillation methods are within the uncertainty of this evaluation.

\section{Full Results}
\label{app:full_results}

\subsection{Intermediate post-training results}
\label{app:full_posttrain_results}
The \olmotwo{} 1B post-training pipeline consists of SFT, DPO, and two stages of RLVR. We reported final post-training results in \cref{tab:post_training_rlvr2_results}; intermediate results for SFT, DPO, and RLVR1 are in \cref{tab:post_training_intermediate_results}.

\begin{table}[!t]
\centering
\renewcommand{\arraystretch}{1.3}
\caption{
\textbf{Per-task results after intermediate post-training stages.}
The NTP baseline is duplicated because it serves as the common reference for both teacher settings.
\textbf{Bold} denotes the best result within each teacher block.
$^{*}$ indicates a statistically significant improvement over the strongest competing baseline ($p<0.05$, paired bootstrap).
Benchmark names are abbreviated for space; see \cref{tab:evaluation_tasks} for full task names.
}
\resizebox{\linewidth}{!}{%
\begin{tabular}{ll|cccccc|ccc|cccccc|c}
\toprule
&
&
\multicolumn{6}{c|}{Reasoning}
&
\multicolumn{3}{c|}{Factual Recall}
&
\multicolumn{6}{c|}{Knowledge \& Commonsense}
&
\multicolumn{1}{c}{Inst.} \\
\cmidrule(lr){3-8}
\cmidrule(lr){9-11}
\cmidrule(lr){12-17}
\cmidrule(l){18-18}
$T$ & Method
& GSM8K & GSM-S & GSM+ & BBH & DROP & MATH
& TQA & NQ & SQA
& MMLU & MMLU-P & ARC-C & OBQA & Wino & AGI
& IFE \\
\midrule
\multicolumn{18}{c}{\textbf{After SFT}} \\
\midrule

N/A & NTP & 45.2 & 28.2 & 23.3 & 30.2 & 33.8 & 6.8 & \textbf{56.3} & 21.7 & 8.2 & 39.2 & 16.3 & 51.1 & 51.4 & 51.5 & 33.2 & 45.8 \\

7B & FKD & 54.7 & 34.2 & 30.7 & 31.3 & 39.1 & 9.4 & 54.1 & 23.3 & 8.2 & 49.1 & 18.5 & \textbf{61.7} & 59.4 & 51.9 & 39.7 & \textbf{46.0} \\

7B & RKD & 57.7 & 35.5 & 32.0 & 32.1 & 39.9 & 11.2 & 53.2 & 22.4 & 8.4 & 48.9 & 18.2 & 60.9 & 60.2 & 52.2 & 39.2 & 45.8 \\

7B & TRKD & 49.9 & 31.1 & 26.9 & 31.5 & 37.1 & 7.6 & 53.7 & 23.1 & 8.4 & 46.5 & 17.4 & 57.8 & 56.0 & 51.4 & 37.5 & 45.3 \\

7B & \textbf{SD} & \textbf{63.7}$^{*}$ & \textbf{42.7}$^{*}$ & \textbf{36.7}$^{*}$ & \textbf{33.5}$^{*}$ & \textbf{48.9}$^{*}$ & \textbf{12.8} & 55.1 & \textbf{24.5}$^{*}$ & \textbf{8.5} & \textbf{50.3}$^{*}$ & \textbf{19.4}$^{*}$ & 61.3 & \textbf{61.4} & \textbf{56.1}$^{*}$ & \textbf{40.6} & 45.5 \\

\midrule[0.8pt]

N/A & NTP & 45.2 & 28.2 & 23.3 & 30.2 & 33.8 & 6.8 & \textbf{56.3} & 21.7 & 8.2 & 39.2 & 16.3 & 51.1 & 51.4 & 51.5 & 33.2 & \textbf{45.8} \\

13B & FKD & 54.1 & 32.5 & 28.0 & 31.8 & 36.3 & 7.6 & 54.9 & 23.5 & 7.9 & 45.9 & 17.1 & 55.6 & 57.2 & 51.2 & 37.1 & 44.2 \\

13B & RKD & 56.8 & 36.7 & 31.3 & \textbf{32.1} & 38.9 & \textbf{11.6} & 55.4 & 23.8 & 8.0 & 47.3 & 17.3 & \textbf{57.8}$^{*}$ & 57.4 & \textbf{52.2} & \textbf{38.1} & 43.6 \\

13B & TRKD & 49.0 & 29.8 & 25.4 & 31.6 & 35.2 & 7.6 & 55.4 & 23.6 & \textbf{8.8} & 43.8 & 16.7 & 53.8 & 55.0 & 51.3 & 36.2 & 43.4 \\

13B & \textbf{SD} & \textbf{62.7}$^{*}$ & \textbf{42.5}$^{*}$ & \textbf{34.8}$^{*}$ & 31.6 & \textbf{43.5}$^{*}$ & 10.2 & 54.9 & \textbf{24.1} & 8.5 & \textbf{48.0}$^{*}$ & \textbf{17.7} & 55.1 & \textbf{59.4} & \textbf{52.2} & 38.0 & 43.1 \\
\midrule
\multicolumn{18}{c}{\textbf{After DPO}} \\
\midrule
N/A & NTP & 52.4 & 33.4 & 28.2 & 32.3 & 33.5 & 6.6 & \textbf{55.4} & 21.5 & 7.7 & 41.1 & 16.4 & 50.3 & 50.6 & 51.5 & 34.1 & 59.9 \\

7B & FKD & 60.4 & 35.8 & 33.1 & 33.1 & 39.8 & 7.8 & 53.6 & 23.1 & \textbf{8.3} & 49.2 & 18.5 & 59.2 & 58.0 & 51.6 & 39.7 & \textbf{64.1} \\

7B & RKD & 64.5 & 40.0 & 36.4 & 34.2 & 40.5 & 10.0 & 52.8 & 22.3 & 7.8 & 48.8 & 17.3 & 57.4 & 55.0 & 52.5 & 38.2 & \textbf{64.1} \\

7B & TRKD & 57.2 & 34.9 & 32.3 & 32.4 & 36.6 & 7.2 & 53.1 & 23.2 & 7.9 & 47.0 & 17.9 & 56.7 & 57.2 & 51.9 & 37.5 & 62.5 \\

7B & \textbf{SD} & \textbf{70.9}$^{*}$ & \textbf{50.8}$^{*}$ & \textbf{44.2}$^{*}$ & \textbf{35.5} & \textbf{49.0}$^{*}$ & \textbf{15.0}$^{*}$ & 54.7 & \textbf{23.6} & 8.1 & \textbf{50.0} & \textbf{20.1}$^{*}$ & \textbf{61.2} & \textbf{61.0} & \textbf{55.5}$^{*}$ & \textbf{40.9} & 62.1 \\

\midrule[0.8pt]

N/A & NTP & 52.4 & 33.4 & 28.2 & 32.3 & 33.5 & 6.6 & 55.4 & 21.5 & 7.7 & 41.1 & 16.4 & 50.3 & 50.6 & 51.5 & 34.1 & 59.9 \\

13B & FKD & 61.0 & 34.2 & 31.7 & \textbf{33.4} & 36.9 & 8.2 & 54.6 & 23.2 & 7.7 & 45.9 & 17.4 & 53.2 & 57.2 & 52.3 & 37.4 & \textbf{64.7} \\

13B & RKD & 65.6 & 41.2 & 35.0 & 33.2 & 39.9 & 7.4 & 53.5 & 23.7 & 7.9 & \textbf{47.8} & \textbf{17.9} & \textbf{57.1} & 56.2 & 51.9 & 38.5 & 64.5 \\

13B & TRKD & 53.1 & 34.9 & 30.7 & 32.9 & 35.3 & 6.8 & \textbf{54.9} & 22.4 & \textbf{8.5} & 43.4 & 16.5 & 53.2 & 56.0 & 52.0 & 35.3 & 61.7 \\

13B & \textbf{SD} & \textbf{69.9}$^{*}$ & \textbf{45.7}$^{*}$ & \textbf{39.5}$^{*}$ & 32.2 & \textbf{44.2}$^{*}$ & \textbf{10.0} & 54.5 & \textbf{23.9} & 8.2 & 47.3 & \textbf{17.9} & 56.1 & \textbf{59.2} & \textbf{52.5} & \textbf{38.7} & 61.0 \\
\midrule 
\multicolumn{18}{c}{\textbf{After RLVR1}} \\
\midrule 
N/A & NTP & 69.1 & 45.6 & 40.5 & 32.4 & 33.4 & 11.2 & \textbf{54.5} & 21.7 & 8.0 & 31.1 & 16.9 & 52.3 & 54.2 & 51.1 & 33.2 & 63.6 \\

7B & FKD & 76.7 & 59.4 & 51.9 & 34.4 & 39.2 & 16.8 & 52.8 & \textbf{23.1} & \textbf{8.3} & 41.4 & 18.3 & 59.4 & 56.2 & 51.2 & 38.0 & 67.1 \\

7B & RKD & 77.9 & 54.3 & 52.5 & 34.2 & 40.1 & 15.6 & 52.1 & 21.9 & 7.8 & 45.3 & 16.8 & 57.6 & 50.4 & 51.5 & 36.1 & 59.3 \\

7B & TRKD & 71.0 & 54.6 & 46.8 & 33.0 & 37.4 & 11.4 & 52.8 & 22.7 & 8.0 & 40.2 & 17.8 & 58.9 & 55.6 & 52.0 & 35.9 & 65.2 \\

7B & \textbf{SD} & \textbf{79.3} & \textbf{63.4}$^{*}$ & \textbf{52.8} & \textbf{36.1}$^{*}$ & \textbf{48.8}$^{*}$ & \textbf{17.4} & 54.1 & 23.0 & 8.0 & \textbf{47.2}$^{*}$ & \textbf{19.7}$^{*}$ & \textbf{61.7} & \textbf{59.8} & \textbf{53.0} & \textbf{39.7}$^{*}$ & \textbf{68.8} \\

\midrule[0.8pt]

N/A & NTP & 69.1 & 45.6 & 40.5 & 32.4 & 33.4 & 11.2 & 54.5 & 21.7 & 8.0 & 31.1 & 16.9 & 52.3 & 54.2 & 51.1 & 33.2 & 63.6 \\

13B & FKD & 73.0 & 55.5 & 47.3 & \textbf{33.7} & 36.1 & 14.0 & \textbf{54.7} & 23.3 & 8.3 & 35.0 & 17.2 & 54.0 & 56.6 & \textbf{51.7} & 35.4 & 63.2 \\

13B & RKD & 75.6 & 60.1 & 50.3 & 33.3 & 38.1 & 15.8 & 53.3 & 22.3 & 8.0 & 36.2 & 16.9 & 55.9 & 55.2 & 51.3 & 36.4 & 63.6 \\

13B & TRKD & 72.4 & 53.3 & 47.6 & 32.7 & 35.0 & 10.4 & \textbf{54.7} & 22.2 & \textbf{8.6} & 36.6 & 16.6 & 54.1 & 54.6 & \textbf{51.7} & 34.5 & 67.1 \\

13B & \textbf{SD} & \textbf{78.8}$^{*}$ & \textbf{65.6}$^{*}$ & \textbf{52.5}$^{*}$ & 33.5 & \textbf{43.3}$^{*}$ & \textbf{17.8} & 54.3 & \textbf{24.0} & 8.2 & \textbf{38.7}$^{*}$ & \textbf{17.7} & \textbf{56.2} & \textbf{57.8} & \textbf{51.7} & \textbf{37.7} & \textbf{68.9} \\
\bottomrule
\end{tabular}}
\label{tab:post_training_intermediate_results}
\end{table}

\subsection{Ablation mid-training results}
\label{app:ablation_full_results}
We provide per-task results for the mid-training ablations to \ours{} in \cref{tab:ablation_results_pertask} (macro-averages are in \cref{tab:ablation_results}).
\begin{table*}[t!]
\centering
\setlength{\tabcolsep}{3pt}
\renewcommand{\arraystretch}{1.15}
\caption{
\textbf{Per-task mid-training ablations to \ours{} (SD), using \olmotwo{} 7B Instruct as the teacher.}
Macro-averaged results appear in \cref{tab:ablation_results}.
}
\setlength{\tabcolsep}{2pt}
\resizebox{\linewidth}{!}{%
\begin{tabular}{l|cccccc|ccc|cccccc}
\toprule
 &
\multicolumn{6}{c|}{Reasoning} &
\multicolumn{3}{c|}{Factual Recall} &
\multicolumn{6}{c}{Knowledge \& Commonsense} \\
\cmidrule(lr){2-7}\cmidrule(lr){8-10}\cmidrule(lr){11-16}
Method & GSM8K & GSM-S & GSM+ & BBH & DROP & MATH & TQA & NQ & SQA & MMLU & MMLU-P & ARC-C & OBQA & Wino & AGI \\
\midrule

\multicolumn{16}{l}{\textbf{Distillation Objective}}\\
\cmidrule(lr){1-16}

\ours{}$_{\mathrm{FKL}}$
& 66.7 & 52.0 & 43.7 & 31.2 & 46.4 & 11.0 & 54.8 & 24.3 & 8.2 & 50.5 & 19.1 & 63.7 & 62.6 & 52.2 & 39.6 \\
\midrule 
\multicolumn{16}{l}{\textbf{Routing Policy}}\\
\cmidrule(lr){1-16}
\textsc{Teacher-Correct Routing}
& 63.8 & 49.5 & 41.7 & 31.8 & 45.2 & 9.8 & 44.5 & 19.4 & 8.7 & 50.6 & 19.3 & 63.2 & 64.6 & 52.0 & 39.9 \\

\textsc{Random Routing}
& 61.5 & 45.5 & 38.6 & 32.0 & 40.6 & 10.8 & 53.2 & 23.9 & 8.5 & 49.6 & 17.8 & 61.6 & 62.6 & 52.6 & 39.5 \\
\textsc{Oracle Domain Routing}
& 61.0 & 45.6 & 38.5 & 31.6 & 40.1 & 8.0 & 53.0 & 22.6 & 8.3 & 49.4 & 18.8 & 61.1 & 61.4 & 52.6 & 38.9 \\

\midrule
\multicolumn{16}{l}{\textbf{Supervision Objective}}\\
\cmidrule(lr){1-16}

\textsc{Always CE}
& 69.1 & 55.2 & 46.5 & 33.3 & 50.1 & 12.0 & 54.3 & 24.5 & 9.4 & 51.1 & 19.9 & 63.7 & 63.6 & 52.7 & 41.1 \\
\textsc{Teacher Top-1}
& 62.2 & 46.0 & 40.1 & 30.7 & 42.1 & 8.8 & 57.9 & 24.7 & 9.3 & 48.8 & 18.0 & 59.8 & 62.0 & 51.0 & 39.2 \\
\bottomrule
\end{tabular}}
\label{tab:ablation_results_pertask}
\end{table*}

\end{document}